\documentclass{sig-alternate}
\usepackage{url}
\usepackage{listings}
\usepackage{tikz}
\usetikzlibrary{positioning}
\usetikzlibrary{shapes.geometric}
\usetikzlibrary{snakes}
\usetikzlibrary{arrows}
\definecolor{dkgreen}{rgb}{0,0.6,0}
\definecolor{gray}{rgb}{0.5,0.5,0.5}
\definecolor{mauve}{rgb}{0.58,0,0.82}

\usepackage[linesnumbered]{algorithm2e}
\usepackage{hhline}
 \usepackage{ulem}
\usepackage{longtable}
\usepackage{multirow}
\usepackage{pgfplots}

\begin{document}

\conferenceinfo{ESEC/FSE'15}{August 31-September 4, 2015, Bergamo, Italy.}
\CopyrightYear{2015} 

\title{Exploration of the scalability of LocFaults approach for error localization with While-loops programs}
%
%
%
%
%

\numberofauthors{1} 
%
\author{
%
%
\alignauthor
Mohammed Bekkouche\\
       \affaddr{University of Nice-Sophia}\\
       \affaddr{Antipolis, I3S/CNRS}\\
       \affaddr{BP 121, 06903 Sophia}\\
       \affaddr{Antipolis Cedex, France}\\
       \email{bekkouch@i3s.unice.fr}
}
\additionalauthors{Additional authors: John Smith (The Th{\o}rv{\"a}ld Group,
email: {\texttt{jsmith@affiliation.org}}) and Julius P.~Kumquat
(The Kumquat Consortium, email: {\texttt{jpkumquat@consortium.net}}).}
\date{30 July 1999}

\maketitle
\begin{abstract}
A model checker can produce a trace of counterexample, for an erroneous program, which is often long and difficult to understand. In general, the part about the loops is the largest among the instructions in this trace. This makes the location of errors in loops critical, to analyze errors in the overall program. In this paper, we explore the scalability capabilities of {\tt LocFaults}, our error localization approach exploiting paths of CFG(Control Flow Graph) from a counterexample to calculate the MCDs (Minimal Correction Deviations), and MCSs (Minimal Correction Subsets) from each found MCD. We present the times of our approach on programs with \textit{While}-loops unfolded $b$ times, and a number of deviated conditions ranging from $0$ to $n$. Our preliminary results show that the times of our approach, constraint-based and flow-driven, are better compared to {\tt BugAssist} which is based on SAT and transforms the entire program to a Boolean formula, and further the information provided by {\tt LocFaults} is more expressive for the user.
\end{abstract}

\category{D.3.3}{Language Constructs and features}{Constraints}
\category{D.2.5}{Testing and Debugging}{Debugging aids, Diagnostics, Error handling and recovery}

\terms{Verification, Algorithms, Experimentation
}

\keywords{Error localization, LocFaults, BugAssist, Off-by-one bug, Minimal Correction Deviations, Minimal Correction Subsets} 

\section{Introduction}
Errors are inevitable in a program, they can harm proper operation and have extremely serious financial consequences. Thus it poses a threat to human well-being~\cite{listofsoftwarebugs}. This link~\cite{bugstories} cites recent stories of software bugs. Consequently, the debugging process (detection, localization and correction of errors) is essential. The location of errors is the step that costs the most. It consists of identifying the exact locations of suspicious instructions~\cite{wong2009survey} to help the user to understand why the program failed, which facilitates him in the task of error correction. Indeed, when a program P is not conformed with its specification (P contains errors), a model checker can produce a trace of a counterexample, which is often long and difficult to understand even for experienced programmers. To solve this problem, we have proposed an approach~\cite{bekkouche2015locfaults} (named {\tt LocFaults}) based on constraints that explores the paths of CFG (Control Flow Graph) of the program from the counterexample, to calculate the minimal subsets to restore the program's compliance with its postcondition. Ensuring that our method is highly scalable to meet the enormous complexity of software systems is an important criterion for its quality~\cite{d2008survey}.

Different statistical approaches for error localization have been proposed; e.g.:  {\tt Tarantula}~\cite{jones2002visualization}~\cite{jones2005empirical}, {\tt Ochiai}~\cite{abreu2007accuracy}, {\tt AMPLE}~\cite{abreu2007accuracy}, {\tt Pinpoint}~\cite{chen2002pinpoint}. The most famous is {\tt Tarantula}, which uses different metrics to calculate the degree of suspicion of each instruction in the program while running a battery of tests. The weakness of these approaches is that they require a lot of test cases, while our approach uses one counterexample. Another critical point in statistical approaches is that they require an oracle to decide if the result of a test case is correct or not. To overcome this problem, we consider the framework of Bounded Model Checking (BMC) which only requires a postcondition or assertion to check.

The idea of our approach is to reduce the problem of error localization to the one which is to compute a minimal set which explains why a CSP (Constraint Satisfaction Problem) is infeasible. The CSP represents the union of constraints of the counterexample, the program, and the assertion or the postcondition violated. The calculated set can be a MCS (Minimal Correction Subset) or a MUS (Minimal Unsatisfiable Subset). In general, test the feasibility of a CSP over a finite domaine is a NP-complete problem (intractable)\footnote{If this problem could be solved in polynomial time, then all NP-complete problems would be too.}, one of the most difficult NP problems. This means, explaining the infeasibility in a CSP is as hard or more (it can be classified as NP-hard problem). {\tt BugAssist}~\cite{jose2011cause}~\cite{jose2011bug} is a BMC method of error localization using a Max-SAT solver to calculate the merger of MCSs of the Boolean formula of the entire program with the counterexample. It becomes inefficient for large programs. {\tt LocFaults} also works from a counterexample to calculate MCSs. 

In this paper, we explore the scalability of {\tt LocFaults} on programs with \textit{While}-loops unfolded $b$ times, and a number of deviated conditions ranging from $0$ to $3$. 

The contribution of our approach against BugAssist can be summarized in the following points:
\begin{itemize}
\item[*] We do not transform the entire program in a system of constraints, but we use the CFG of the program to collect the constraints of the path of counterexample and paths derivatives thereof, assuming that at most $k$ conditionals may contain errors. We calculate MCSs only on the path of counterexample and paths that correct the program;
\item[*] We do not translate the program instructions into a SAT formula, instead  numerical constraints that will be handled by constraint solvers;
\item[*] We do not use MaxSAT solvers as black boxes, instead a generic algorithm to calculate MCSs by the use of a constraint solver;
\item[*] We limit the size of the generated MCSs and the number of deviated conditions;
\item[*] We can work together more solvers during the localization process and take the most efficient according to the category of CSP constructed. For example, if the CSP of the path detected is of type linear over integers, we use a MIP (Mixed Integer Programming) solver; if it is nonlinear, we use a CP (Constraint Programming) solver and/or as well as MINLP (Mixed Integer Nonlinear Programming).
\end{itemize}
Our practical experience has shown that all these restrictions and distinctions enable {\tt LocFaults} to be faster and more expressive.

The paper is organized as follows. Section 2 introduces the definition of MUS and MCS. In Section 3, we define the problem $\leq k$-MCD. We explain a paper contribution for the treatment of erroneous loops, including the \textit{Off-by-one} bug, in Section 4. A brief description of our {\tt LocFaults} algorithm is provided in Section 5. The experimental evaluation is presented in Section 6. Section 7 talks about the conclusion and future work.

\section{Definitions}
In this section, we introduce the definition of an IIS/MUS and MCS.
\paragraph{CSP} A CSP (Constraint Satisfaction Problem) $P$ is defined as a triple $<X,D,C>$, where:
\begin{itemize}
\item[*] $X$ a set of $n$ variables $x_{1}, x_{2}, ..., x_{n}$.
\item[*] $D$ the tuple $<D_{x_{1}}, D_{x_{2}}, ..., D_{x_{n}}>$. The set $D_{x_{i}}$ contains the values of the variable $x_{i}$.   
\item[*] $C$=$\{c_{1}, c_{2}, ..., c_{n}\}$ is the set of constraints.
\end{itemize}
A \textit{solution} for $P$ is an instantiation of the variables $\cal{I}$ $\in$ $D$ that satisfies all the constraints in $C$. $P$ is infeasible if it has no solutions. A sub-set of constraints $C'$ in $C$ is also said infeasible for the same reason except that it is limited to the constraints in $C'$.\newline
We denote as:
\begin{itemize}
\item $Sol(<X,C',D>)=\emptyset$, to specify that $C'$ has no solutions, so it is unfeasible.
\item $Sol(<X,C',D>) \neq \emptyset$, to specify that $C'$ has at least one solution, so it is feasible.
\end{itemize}

We say that $P$ is \textit{linear} and denote LP (Linear Program) iff all constraints in $C$ are linear equations/inequalities, it is \textit{continuous} if the domain all variables is real. If at least one of the variables in $X$ is integer or binary (Special cases of an integer), and the constraints are linear, $P$ is called a program \textit{linear mixed} MIP (Mixed-integer linear program).
If the constraints are nonlinear, we say that $P$ is a program \textit{nonlinear} NLP (NonLinear Program).

Let $P=<X,D,C>$ an infeasible $CSP$, we define for $P$:
\paragraph{IS}
An IS (Inconsistent Set) is an infeasible subset of constraints in the constraint set infeasible $C$.
$C'$ is an IS iff:
\begin{itemize}
\item[*] $C'$ $\subseteq$ $C$. 
\item[*] $Sol(<X,C',D>)=\emptyset$.
\end{itemize}

\paragraph{IIS or MUS}
An IIS (Irreducible Inconsistent Set) or MUS (Minimal Unsatisfiable Subset) is an infeasible subset of constraints of $C$, and all its strict subsets are feasible.
$C'$ is an IIS iff :
\begin{itemize}
\item[*] $C'$ is an IS.
\item[*] $\forall$ $C''$ $\subset$ $C'$.$Sol(<X,C'',D>)\neq\emptyset$, (each of its parts contributes to the infeasibility), $C'$ is called irreducible.
\end{itemize}

\paragraph{MCS}
$C'$ is a MCS(Minimal Correction Set) iff :
\begin{itemize}
\item[*] $C'$ $\subseteq$ $C$.
\item[*] $Sol(<X,C \backslash C',D>)\neq\emptyset$.
\item[*] $\nexists$ $C''$ $\subset$ $C'$ such as $Sol(<X,C \backslash C'',D>)\neq\emptyset$.
\end{itemize}

\section{The problem $\leq k$-MCD}
Given an erroneous program modeled in CFG\footnote{We use Dynamic Single Assignment (DSA) form~\cite{barnett2005weakest} transformation that ensures that each variable is assigned only once on each path of the CFG.} $G=(C,A,E)$: $C$ is the set of conditional nodes; $A$ is the set of assignment blocks; $E$ is the set of arcs, and a counterexample. A MCD (\textit{Minimal Correction Deviation}) is a set $D$ $\subseteq$ $C$ such as the propagation of the counterexample on all the instructions of $G$ from the root, while having denied each condition\footnote{The condition is denied to take the branch opposite to that where we had to go.} in $D$, allows the output to satisfy the postcondition. It is called minimal (or irreducible) in the sense that no element can be removed from $D$ without losing this property. In other words, $D$ is a minimal program correctness in the set of conditions. The size of minimal deviation is its cardinal. The problem \textit{$\leq k$-MCD} is to find all MCDs of size smaller or equal to $k$.

For example, the CFG of the program AbsMinus (see fig. 2) has one minimal size deviation $1$ for the counterexample $\{i=0,j=1\}$. Certainly, the deviation \{$i_0 \leq j_0$,$k_1=1 \land i_0 \neq j_0$\} corrects the program, but it is not minimal; only one minimal correction deviation for this program is \{$k_1=1 \land i_0 \neq j_0$\}.
 
\begin{figure}[!h]
\begin{minipage}{1cm}
\end{minipage} \hfill
\begin{minipage}[b]{.45\linewidth}
\begin{lstlisting}{}
class AbsMinus {	
/*@ ensures 
  @ ((i<j)==>(\result==j-i))&& 
  @ ((i>=j)==>(\result==i-j));*/
  int AbsMinus (int i, int j){
   int result;
   int k = 0;
	 if (i <= j) {
	  k = k+2;//error: should be k=k+1
	 }
   if (k == 1 && i != j) {
	  result = j-i; 		
	 }  
	 else {
	  result = i-j;
	 }
  }
}
\end{lstlisting}
\vspace{-1cm}
\caption{{\scriptsize The program AbsMinus}}
\label{AbsMinus}
\end{minipage}
\begin{minipage}[b]{.45\linewidth}
\begin{tiny}  
\begin{tikzpicture}[auto,node distance=2cm,/.style={font=\sffamily\footnotesize}]
\tikzstyle{decision} = [scale=0.6,diamond, draw=black, thick,text width=3em, text badly centered, inner sep=1pt]
\tikzstyle{block} = [scale=0.6,rectangle, draw=black, thick, 
text width=5.5em, text centered, rounded corners, minimum height=0.8em,text height=0.5em]
\tikzstyle{bloc} = [scale=0.6,rectangle, draw=black, thick, text centered, minimum height=0.8em,text height=0.5em]
\tikzstyle{line} = [scale=0.6,draw, thick, -latex',shorten >=1pt];
\tikzstyle{cloud} = [scale=0.6,thick, ellipse,draw=black, minimum height=2em];
\matrix [column sep=0.5mm,row sep=2mm,ampersand replacement=\&]
{
\& \node [block] (4) {$k_0=0$}; \& \\
\& \node [decision] (5) {$i_0$ $\leq$ $j_0$}; \& \\
\node [block] (3) {$k_1=k_0+2$};
\& \node[draw=red] (11) {Error}; \& \node [block] (6) {$k_1=k_0$}; \\
\& \node [decision,text width=6.3em] (7) {$k_1=1 \land i_0!=j_0$}; \&  \\ 
\node [block] (8) {$r_1=j_0-i_0$};
\&  \& \node [block] (9) {$r_1=i_0-j_0$}; \\
\& \node [bloc] (10) {POST:$\{r_1==|i-j|\}$}; \& \\
};
\tikzstyle{every path}=[line,distance=1cm]
\path (4) -- (5);
\path (5.west) -- node [left] {If} (3.north);
\path (5.east) -- node [right] {Else} (6.north);
\path (3.south) -- (7);
\path (6.south) -- (7);
\path (7.west) -- node [left] {If} (8.north);
\path (7.east) -- node [right] {Else} (9.north);
\path (8.south) -- (10.west);
\path (9.south) -- (10.east);
\path[snake=zigzag] (11.west) -- (3.east);
\end{tikzpicture} 
\vspace{-0.5cm}
\label{cfgAbsMinus}
\caption{{\scriptsize The CFG in DSA of AbsMinus}}
\end{tiny}
\end{minipage}
\end{figure}

\begin{figure}[!h]
\begin{minipage}{2cm}
\end{minipage} \hfill
\begin{minipage}{4cm}
\begin{tiny}
\begin{tikzpicture}[auto,node distance=2cm,/.style={font=\sffamily\footnotesize}]
\tikzstyle{decision} = [scale=0.6,diamond, draw=black, thick,text width=3em, text badly centered, inner sep=1pt]
\tikzstyle{block} = [scale=0.6,rectangle, draw=black, thick, 
text width=5.5em, text centered, rounded corners, minimum height=0.8em,text height=0.5em]
\tikzstyle{bloc} = [scale=0.6,rectangle, draw=black, thick, text centered, minimum height=0.8em,text height=0.5em]
\tikzstyle{line} = [scale=0.6,draw, thick, -latex',shorten >=1pt];
\tikzstyle{cloud} = [scale=0.6,thick, ellipse,draw=black, minimum height=2em];
\matrix [column sep=0.5mm,row sep=2mm,ampersand replacement=\&]
{
\& \node [bloc,draw=red] (1) {$\{(i_0==0)$ $\land$ $(j_0==1)\}$}; \& \\
\& \node [block,draw=red] (4) {$k_0=0$}; \& \\
\& \node [decision,draw=red] (5) {$i_0$ $\leq$ $j_0$}; \& \\
\node [block,draw=red] (3) {$k_1=k_0+2$};
\&  \& \node [block] (6) {$k_1=k_0$}; \\
\& \node [decision,text width=6.3em,draw=red] (7) {$k_1=1 \land i_0!=j_0$}; \&s  \\ 
\node [block] (8) {$r_1=j_0-i_0$};
\&  \& \node [block,draw=red] (9) {$r_1=i_0-j_0$}; \\
\& \node [bloc,draw=red] (10) {$\{r_1==|i-j|\}$}; \& \\
};
\tikzstyle{every path}=[line,distance=1cm]
\path[draw=red] (1) -- (4);
\path[draw=red] (4) -- (5);
\path[draw=red] (5.west) -- node [left] {If} (3.north);
\path (5.east) -- node [right] {Else} (6.north);
\path[draw=red] (3.south) -- (7);
\path (6.south) -- (7);
\path (7.west) -- node [left] {If} (8.north);
\path[draw=red] (7.east) -- node [right] {Else} (9.north);
\path (8.south) -- (10.west);
\path[draw=red] (9.south) -- (10.east);
\end{tikzpicture}
 \vspace{-0.5cm}
\caption{{\scriptsize The path of the counterexample}} 
\label{cfgAbsMinus1} 
\end{tiny}
\end{minipage} \hfill
\begin{minipage}[c]{4cm}
\begin{tiny}
\begin{tikzpicture}[auto,node distance=2cm,/.style={font=\sffamily\footnotesize}]
\tikzstyle{decision} = [scale=0.6,diamond, draw=black, thick,text width=3em, text badly centered, inner sep=1pt]
\tikzstyle{block} = [scale=0.6,rectangle, draw=black, thick, 
text width=5.5em, text centered, rounded corners, minimum height=0.8em,text height=0.5em]
\tikzstyle{bloc} = [scale=0.6,rectangle, draw=black, thick, text centered, minimum height=0.8em,text height=0.5em]
\tikzstyle{line} = [scale=0.6,draw, thick, -latex',shorten >=1pt];
\tikzstyle{cloud} = [scale=0.6,thick, ellipse,draw=black, minimum height=2em];
\matrix [column sep=0.5mm,row sep=2mm,ampersand replacement=\&]
{
\& \node [bloc,draw=red] (1) {$\{(i_0==0)$ $\land$ $(j_0==1)\}$}; \& \\
\& \node [block,draw=red] (4) {$k_0=0$}; \& \\
\& \node [decision,draw=red,fill=red!5] (5) {$i_0$ $\leq$ $j_0$}; \& \\
\node [block] (3) {$k_1=k_0+2$};
\&  \& \node [block,draw=red] (6) {$k_1=k_0$}; \\
\& \node [decision,text width=6.3em,draw=red] (7) {$k_1=1 \land i_0!=j_0$}; \&s  \\ 
\node [block] (8) {$r_1=j_0-i_0$};
\&  \& \node [block,draw=red] (9) {$r_1=i_0-j_0$}; \\
\& \node [bloc,draw=red] (10) {$\{r_1==|i-j|\}$ is UNSAT}; \& \\
};
\tikzstyle{every path}=[line,distance=1cm]
\path[draw=red] (1) -- (4);
\path[draw=red] (4) -- (5);
\path (5.west) -- node [left] {If} (3.north);
\path[draw=red] (5.east) -- node [right] {Else} (6.north);
\path (3.south) -- (7);
\path[draw=red] (6.south) -- (7);
\path (7.west) -- node [left] {If} (8.north);
\path[draw=red] (7.east) -- node [right] {Else} (9.north);
\path (8.south) -- (10.west);
\path[draw=red] (9.south) -- (10.east);
\end{tikzpicture}
\vspace{-0.5cm}
\caption{{\scriptsize The path obtained by deviating the condition $i_0$ $\leq$ $j_0$}} 
\label{cfgAbsMinus2}  
\end{tiny}
\end{minipage} \hfill
\begin{minipage}{2cm}
\end{minipage} \hfill
\end{figure}

\begin{figure}[!h]
\begin{minipage}{2cm}
\end{minipage} \hfill
\begin{minipage}[c]{4cm}
\begin{tiny}
\begin{tikzpicture}[auto,node distance=2cm,/.style={font=\sffamily\footnotesize}]
\tikzstyle{decision} = [scale=0.6,diamond, draw=black, thick,text width=3em, text badly centered, inner sep=1pt]
\tikzstyle{block} = [scale=0.6,rectangle, draw=black, thick, 
text width=5.5em, text centered, rounded corners, minimum height=0.8em,text height=0.5em]
\tikzstyle{bloc} = [scale=0.6,rectangle, draw=black, thick, text centered, minimum height=0.8em,text height=0.5em]
\tikzstyle{line} = [scale=0.6,draw, thick, -latex',shorten >=1pt];
\tikzstyle{cloud} = [scale=0.6,thick, ellipse,draw=black, minimum height=2em];
\matrix [column sep=0.5mm,row sep=2mm,ampersand replacement=\&]
{
\& \node [bloc,draw=red] (1) {$\{(i_0==0)$ $\land$ $(j_0==1)\}$}; \& \\
\& \node [block,draw=red] (4) {$k_0=0$}; \& \\
\& \node [decision,draw=red] (5) {$i_0$ $\leq$ $j_0$}; \& \\
\node [block,draw=red] (3) {$k_1=k_0+2$};
\&  \& \node [block] (6) {$k_1=k_0$}; \\
\& \node [decision,text width=6.3em,draw=red,fill=red!5] (7) {$k_1=1 \land i_0!=j_0$}; \&s  \\ 
\node [block,draw=red] (8) {$r_1=j_0-i_0$};
\&  \& \node [block] (9) {$r_1=i_0-j_0$}; \\
\& \node [bloc,draw=red] (10) {\textbf{$\{r_1==|i-j|\}$ {is SAT}}}; \& \\
};
\tikzstyle{every path}=[line,distance=1cm]
\path[draw=red] (1) -- (4);
\path[draw=red] (4) -- (5);
\path[draw=red] (5.west) -- node [left] {If} (3.north);
\path (5.east) -- node [right] {Else} (6.north);
\path[draw=red] (3.south) -- (7);
\path (6.south) -- (7);
\path[draw=red] (7.west) -- node [left] {If} (8.north);
\path (7.east) -- node [right] {Else} (9.north);
\path[draw=red] (8.south) -- (10.west);
\path (9.south) -- (10.east);
\end{tikzpicture}
\vspace{-0.5cm}
\caption{{\scriptsize The path by deviating the condition $k_1=1 \land i_0!=j_0$}} 
\label{cfgAbsMinus3}
\end{tiny} 
\end{minipage} \hfill
\begin{minipage}{4cm}
\begin{tiny}
\begin{tikzpicture}[auto,node distance=2cm,/.style={font=\sffamily\footnotesize}]
\tikzstyle{decision} = [scale=0.6,diamond, draw=black, thick,text width=3em, text badly centered, inner sep=1pt]
\tikzstyle{block} = [scale=0.6,rectangle, draw=black, thick, 
text width=5.5em, text centered, rounded corners, minimum height=0.8em,text height=0.5em]
\tikzstyle{bloc} = [scale=0.6,rectangle, draw=black, thick, text centered, minimum height=0.8em,text height=0.5em]
\tikzstyle{line} = [scale=0.6,draw, thick, -latex',shorten >=1pt];
\tikzstyle{cloud} = [scale=0.6,thick, ellipse,draw=black, minimum height=2em];
\matrix [column sep=0.5mm,row sep=2mm,ampersand replacement=\&]
{
\& \node [bloc,draw=red] (1) {$\{(i_0==0)$ $\land$ $(j_0==1)\}$}; \& \\
\& \node [block,draw=red] (4) {$k_0=0$}; \& \\
\& \node [decision,draw=red,fill=red!5] (5) {$i_0$ $\leq$ $j_0$}; \& \\
\node [block] (3) {$k_1=k_0+2$};
\&  \& \node [block,draw=red] (6) {$k_1=k_0$}; \\
\& \node [decision,text width=6.3em,draw=red,fill=red!5] (7) {$k_1=1 \land i_0!=j_0$}; \&s  \\ 
\node [block,draw=red] (8) {$r_1=j_0-i_0$};
\&  \& \node [block] (9) {$r_1=i_0-j_0$}; \\
\& \node [bloc,draw=red] (10) {{\textbf{$\{r_1==|i-j|\}$ {is SAT}}}}; \& \\
};
\tikzstyle{every path}=[line,distance=1cm]
\path[draw=red] (1) -- (4);
\path[draw=red] (4) -- (5);
\path (5.west) -- node [left] {If} (3.north);
\path[draw=red] (5.east) -- node [right] {Else} (6.north);
\path (3.south) -- (7);
\path[draw=red] (6.south) -- (7);
\path[draw=red] (7.west) -- node [left] {If} (8.north);
\path (7.east) -- node [right] {Else} (9.north);
\path[draw=red] (8.south) -- (10.west);
\path (9.south) -- (10.east);
\end{tikzpicture}
\vspace{-0.5cm}
\caption{{\scriptsize The path of a non-minimal deviation:$\{i_0 \leq j_0,k_1=1 \land i_0!=j_0\}$}} 
\label{cfgAbsMinus4}
\end{tiny}
\end{minipage}
\begin{minipage}{2cm}
\end{minipage} \hfill
\vspace{-0.5cm}
\end{figure}

\newpage

The table~\ref{prograssLocFaultsAbsMinus} summarizes the progress of {\tt LocFaults} for the program AbsMinus, with at most 2 conditions deviated from the following counterexample $\{i=0,j=1\}$. 

\begin{table}
\begin{scriptsize}
\begin{tabular}{|c|c|c|c|}
  \hline
  \textit{Conditions deviated} & \textit{MCD}  & \textit{MCS} & Figure  \\
  \hline
  \multirow{2}{*}{$\emptyset$} & \multirow{2}{*}{/} & \multirow{2}{*}{$\{r_1=i_0-j_0:15\}$} & \multirow{2}{*}{fig. 3} \\
         &   &     &  \\   
  \hline
  $\{i_0 \leq j_0:8\}$ & Non & / & fig. 4 \\
  \hline
  \multirow{2}{*}{$\{k_1 = 1 \land i_0 != j_0:11\}$} & \multirow{2}{*}{Oui} & $\{k_0=0:7\}$, & \multirow{2}{*}{fig. 5} \\
         &  & $\{k_1=k_0+2:9\}$ &  \\  
  \hline
  $\{i_0 \leq j_0:8,$ & \multirow{2}{*}{Non}   & \multirow{2}{*}{/} & \multirow{2}{*}{fig. 6} \\  
  $ k_1 = 1 \land i_0 != j_0:11\}$  &  &  &  \\  
  \hline
\end{tabular}
\end{scriptsize}
\vspace{-0.2cm}
\caption{The progress of {\tt LocFaults} for the program AbsMinus.}
\label{prograssLocFaultsAbsMinus}
\end{table}

We display the conditions deviated, if they are minimal deviation or non minimal, and the calculated MCSs from the constructed constraint system : see respectively the columns 1, 2 and 3. Column 4 shows the figure illustrating the path explored for each deviation. In the first and the third column we show in addition of the instruction, its line in the program. For example, the first line in the table shows that there is a single MCS found ($\{r_1=i_0-j_0:15\}$) on the path of the counterexample.

\section{Error localization in loops}
As part of Bounded Model Checking (BMC) for programs, unfolding can be applied to the entire program or it can be applied to loops separately~\cite{d2008survey}. Our algorithm LocFaults~\cite{bekkouche2014approche}~\cite{bekkouche2015locfaults} for error localization is placed in the second approach; that is to say, we use a bound $b$ to unfold loops by replacing them with conditional statements nested of depth $b$. Consider for instance the program Minimum (see fig. 7), containing a single loop, that calculates the minimum in an array of integers. The effect on control flow graph of the program  Minimum before and after unfolding is illustrated in Figures 7 and 8 respectively. The \textit{While}-loop is unfolded 3 times, as 3 is the number of iterations needed for the loop to calculate the minimum value in an array of size 4 in the worst case.

{\tt LocFaults} takes as input the CFG of the erroneous program, $CE$ a counterexample, $b_ {mcd}$: a bound on the number of deviated conditions, $b_ {mcs}$: a bound on the size of MCSs calculated. It allows to explore the CFG in depth by diverting at most $b_ {mcd}$ conditions from the path of the counterexample:
\begin{itemize}
\item[*] It propagates $CE$ on the CFG until the postcondition. Then it calculates the MCSs on the CSP of the path generated to locate errors on the  path of counterexample. 
\item[*] It seeks to enumerate the sets $\leq b_{mcd}$-MCD. For each found MCD, it calculates the MCSs on the path that arrives at the last deviated condition and allows to take the path of the deviation.  
\end{itemize} 
 
Among the most common errors associated with loops according to~\cite{debugloopswp}, the \textit{Off-by-one} bug,  i.e. loops that iterate one too many or one too few times. This may be due to improper initialization of the loop control variables, or an erroneous condition of the loop. The program \textit{Minimum} presents a case of this type of error. It is erroneous because of its loop \textit{While}, the falsified instruction is on the condition of the loop (line 9): the correct condition should be $(i<tab.length)$ ($tab.length$ is the number of elements of the table $tab$). From the following counterexample $\{tab[0]=3, tab[1]=2, tab[2]=1, tab[3]=0\}$, we illustrated in Figure 8 the initial faulty path (see the colorful path in red) and the deviation for which the postcondition is satisfiable (the deviation and the path above the deviated condition are shown in green).

\begin{figure}[h]
\begin{minipage}{2cm}
\end{minipage} \hfill
\begin{minipage}[b]{.45\linewidth}
\begin{lstlisting}{}
class Minimum {
/* The minimum in an array of n integers*/
/*@ ensures 
  @ (\forall int k;(k >= 0 && k < tab.length);tab[k] >= min);
  @*/
  int Minimum (int[] tab) {
    int min=tab[0];
    int i = 1;
    while (i<tab.length-1) { /*error, the condition should be (i<tab.length)*/
      if (tab[i]<=min){
        min=tab[i];
      }
     	i = i+1;
    }
    return min;
  }
}
\end{lstlisting}

\end{minipage} \hfill
\begin{minipage}[b]{.52\linewidth}

\begin{scriptsize}
\begin{tikzpicture}[auto,node distance=2cm,/.style={font=\sffamily\footnotesize}]
\tikzstyle{decision} = [scale=0.6,diamond, draw=black, thick,text width=3em, text badly centered, inner sep=1pt]
\tikzstyle{block} = [scale=0.6,rectangle, draw=black, thick, 
text width=6em, text centered, rounded corners, minimum height=0.8em,text height=0.5em]
\tikzstyle{bloc} = [scale=0.6,rectangle, draw=black, thick, text centered, minimum height=0.8em,text height=0.5em,minimum width=3cm]
\tikzstyle{line} = [scale=0.6,draw, thick, -latex',shorten >=1pt];
\tikzstyle{cloud} = [scale=0.6,thick, ellipse,draw=black, minimum height=2em];
\matrix [column sep=0.5mm,row sep=2mm,ampersand replacement=\&]
{
\& \node [block] (5) {$min=tab[0]$  $i = 1$}; \& \\

\&  \&  \\

\& \node [decision,text width=6.3em] (6) {$i<tab.length-1$}; \& \\

\& \node [decision,text width=6.3em] (7) {$tab[i] \leq min$}; \& \\

\&  \&  \\

\& \node [block] (8) {$min=tab[i]$}; \&  \\

\&  \&  \\

\&  \&  \\

\& \node [block] (9) {$i=i+1$}; \& \\

\& \node [bloc] (18) {Postcondition}; \& \\
};
\tikzstyle{every path}=[line,distance=0.5cm]
\path (5) -- (6);
\path (6.west) -- node [left] {If} (7);
\path (7.west) -- node [left] {If} (8);
\path (8) -- (9.north);
\draw[->] (7.east) .. controls +(right:10mm) and +(right:10mm) .. node [right] {Else} (9.east);
\draw[->] (6.east) .. controls +(right:20mm) and +(right:20mm) .. node [right] {Else} (18.east);
\draw[->] (9.west) .. controls +(left:20mm) and +(left:30mm) .. node [left] {Goto} (6);
\end{tikzpicture}
\end{scriptsize}
\end{minipage}

\label{MinimumCFG}
\vspace{-0.4cm}
\caption{The program Minimum and its normal CFG (non unfolded). The postcondition is $\{\forall\text{ int k};(k\geq0 \land k<tab.length);tab[k] \geq min\}$}
\end{figure}
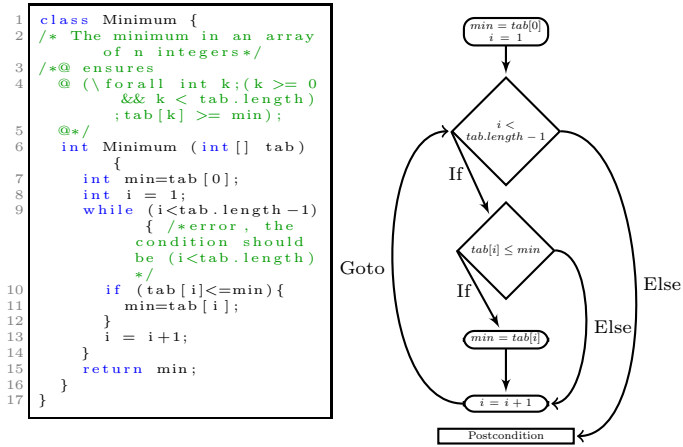

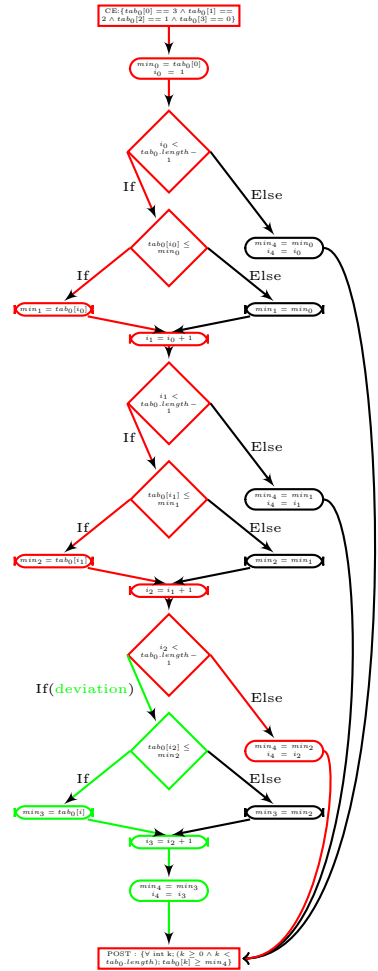
\begin{figure}[h]
\center
\begin{tiny}
\begin{tikzpicture}[scale=0.5,auto,node distance=2cm,/.style={font=\sffamily\footnotesize}]
\tikzstyle{decision} = [scale=0.5,diamond, draw=black, thick,text width=4em, text badly centered, inner sep=1pt]
\tikzstyle{block} = [scale=0.5,rectangle, draw=black, thick, 
text width=8em, text centered, rounded corners, minimum height=0.8em,text height=0.5em];
\tikzstyle{bloc} = [scale=0.5,rectangle, draw=black, thick, 
text width=15em, text centered, minimum height=0.8em,text height=0.5em];
\tikzstyle{line} = [scale=0.5,draw, thick, -latex',shorten >=1pt];
\tikzstyle{cloud} = [scale=0.5,thick, ellipse,draw=black, minimum height=2em];
\matrix [column sep=0.5mm,row sep=2mm,ampersand replacement=\&]
{

\& \node [bloc,draw=red] (22) {CE:$\{tab_0[0]==3 \land tab_0[1]==2 \land tab_0[2]==1 \land tab_0[3]==0\}$}; \& \\

\&  \&  \\

\& \node [block,draw=red] (5) {$min_0=tab_0[0]$  $i_0 = 1$}; \& \\

\&  \&  \\

\& \node [decision,text width=6.3em,draw=red] (6) {$i_0$ $<$ $tab_{0}.length-1$}; \& \\

\& \node [decision,text width=6.3em,draw=red] (7) {$tab_{0}[i_0]$ $\leq$ $min_0$}; \& \node [block] (23) {$min_4=min_0$  $i_4=i_0$};  \\

\node [block,draw=red] (8) {$min_{1}$ $=$ $tab_{0}[i_0]$}; \&  \& \node [block] (19) {$min_{1}$ $=$ $min_{0}$}; \\

\& \node [block,draw=red] (9) {$i_{1}$ $=$ $i_{0}+1$}; \& \\

\& \node [decision,text width=6.3em,draw=red] (10) {$i_1$ $<$ $tab_{0}.length-1$}; \& \\

\& \node [decision,text width=6.3em,draw=red] (11) {$tab_{0}[i_1]$ $\leq$ $min_1$}; \& \node [block] (24) {$min_4=min_1$  $i_4=i_1$}; \\

\node [block,draw=red] (12) {$min_2$ $=$ $tab_{0}[i_1]$}; \&  \& \node [block] (20) {$min_{2}$ $=$ $min_{1}$};  \\

\& \node [block,draw=red] (13) {$i_2$ $=$ $i_1+1$}; \& \\

\& \node [decision,text width=6.3em,draw=red] (14) {$i_2$ $<$ $tab_{0}.length-1$}; \& \\

\& \node [decision,text width=6.3em,draw=green] (15) {$tab_{0}[i_2]$ $\leq$ $min_2$}; \& \node [block,draw=red] (25) {$min_4=min_2$  $i_4=i_2$}; \\

\node [block,draw=green] (16) {$min_3$ $=$ $tab_{0}[i]$}; \&  \& \node [block] (21) {$min_{3}$ $=$ $min_{2}$};  \\

\& \node [block,draw=green] (17) {$i_3$ $=$ $i_2+1$}; \& \\

\&  \& \\

\& \node [block,draw=green] (26) {$min_4=min_3$  $i_4=i_3$}; \& \\

\&  \& \\

\&  \& \\ 

\& \node [bloc,draw=red] (18) {POST : $\{\forall\text{ int k};(k\geq0 \land k<tab_{0}.length);tab_{0}[k] \geq min_4\}$}; \& \\
};
\tikzstyle{every path}=[line,distance=1cm]
\path[draw=red] (22) -- (5);
\path[draw=red] (5) -- (6);
\path[draw=red] (6.west) -- node [left] {If} (7);
\path (6.east) -- node [right] {Else} (23);
\path[draw=red] (7.west) -- node [left] {If} (8);
\path (7.east) -- node [right] {Else} (19);
\path[draw=red] (8) -- (9.north);
\path (19) -- (9.north);
\path[draw=red] (9) -- (10);
\path[draw=red] (10.west) -- node [left] {If} (11);
\path (10.east) -- node [right] {Else} (24);
\path[draw=red] (11.west) -- node [left] {If} (12);
\path (11.east) -- node [right] {Else} (20);
\path[draw=red] (12) -- (13.north);
\path (20) -- (13.north);
\path[draw=red] (13) -- (14);
\path[draw=green] (14.west) -- node [left] {If(\textcolor{green}{deviation})} (15);
\path[draw=red] (14.east) -- node [right] {Else} (25);
\path[draw=green] (15.west) -- node [left] {If} (16);
\path (15.east) -- node [right] {Else} (21);
\path[draw=green] (16) -- (17.north);
\path (21) -- (17.north);
\path[draw=green] (17) -- (26);
\path[draw=green] (26) -- (18);
\draw[->] (23.east) .. controls +(right:35mm) and +(right:95mm) ..  (18.east);
\draw[->] (24.east) .. controls +(right:25mm) and +(right:70mm) ..  (18.east);
\draw[->,draw=red] (25.east) .. controls +(right:10mm) and +(right:45mm) ..  (18.east);
\end{tikzpicture}
\end{tiny}
\label{MinimumCFGDeplie}
\vspace{-0.1cm}
\caption{{\scriptsize Figure showing the CFG in DSA form of the program \textit{Minimum} by unfolding its loop $3$ times, with the path of a counterexample (shown in red) and a deviation satisfying the postcondition (shown in green).}}
\vspace{-0.6cm}
\end{figure}

We show in table~\ref{PathsMcs} erroneous paths generated (column $PATH$) and the MCSs calculated (column $MCSs$) for at most $1$ condition deviated from the conduct of the counterexample. The first line concerns the path of counterexample; the second for the path obtained by deviating the condition $\{i_2 \leq tab_0.length-1\}$. \newline
\begin{table}
\begin{scriptsize}
\begin{tabular}{|c|c|}
\hline
 $PATH$ & MCSs \\ 
\hline
 $\{CE:[tab_0[0]=3 \land tab_0[1]=2 \land tab_0[2]=1$  & \multirow{5}{*}{$\{min_2=tab_0[i_1]\}$}  \\
 $\land  tab_0[3]==0]$, $min_0=tab_0[0]$, $i_0=1$, &  \\
 $min_1=tab_0[i_0]$,$i_1=i_0+1$,$min_2=tab_0[i_1]$, &  \\
 $i_2=i_1+1$,$min_3=min_2$, $i_3=i_2$,  &   \\ 
 $POST:[(tab[0] \geq min_3) \land  (tab[1] \geq min_3)$  &  \\
 $\land (tab[2] \geq min_3) \land (tab[3] \geq min_3)]\}$ &  \\
\hline
 $\{CE:[tab_0[0]=3 \land tab_0[1]=2 \land tab_0[2]=1$  & \multirow{2}{*}{$\{i_0=1\}$,}  \\
 $\land  tab_0[3]==0]$, $min_0=tab_0[0]$, $i_0=1$, & \multirow{2}{*}{$\{i_1=i_0+1\}$,} \\
 $min_1=tab_0[i_0]$,$i_1=i_0+1$,$min_2=tab_0[i_1]$,  &  \multirow{2}{*}{$\{i_2=i_1+1\}$} \\
 $i_2=i_1+1$,$[\neg (i_2 \leq tab_0.length-1)]$  &  \\
\hline
\end{tabular}
\end{scriptsize}
\vspace{-0.1cm}
\caption{Paths and MCSs generated by {\tt LocFaults} for the program \textit{Minimum}.}
\label{PathsMcs}
\end{table}
 
{\tt LocFaults} identifies a single MCS on the path of counterexample that contains the constraint $min_2=tab_0[i_1]$, the instruction of the line $11$ in the second iteration of the loop unfolded. With a deviated condition, the algorithm  suspects the third condition of the unfolded loop $i_2$ $<$ $tab_{0}.length-1$; in other words, we need a new iteration to satisfy the postcondition.

This example shows a case of a program with an incorrect loop: the error is on the stopping criterion, it does not allow the program to iterate until the last element of the array input. LocFaults with its deviation mechanism is able to detect this type of error accurately. It provides the user not only suspicious instructions in the loop not unfolded on the original program, but also information about the iterations where they are in the unfolded loop. This information could be very useful for the programmer to understand the errors in the loop.

\section{Algorithm}
Our goal is to find MCDs of size less than a bound $k$ ; in other words, we try to give a solution to the problem posed above ($\leq k$-MCD). For this, our algorithm (named {\tt LocFaults}) explores in depth the CFG  and generates the paths where at most $k$ conditions are deviated from the conduct of the counterexample.

To improve efficiency, our heuristic solution proceeds incrementally. It successively deviates from $0$ to $k$ conditions and search the MCSs for the corresponding paths. However, if in step $k$ {\tt LocFaults} deviates a condition $c_i$  and that it has corrected the program, it does not explore in step $k'$ with $k'>k$ paths that involve a deviation from the condition $c_i$. For this, we add the cardinality of the found minimum deviation ($k$) as information on the node of $c_i$.
 
We will illustrate with an example of our approach, as seen in the graph in Figure 9. Each circle in the graph represents a conditional node visited by the algorithm. The example does not show the block of assignments because we want to illustrate just how we find the minimal correction deviations of a bounded size as mentioned above. An arc connecting a condition $c_1$ to another $c_2$ illustrates that $c_2$ is reached by the algorithm. There are two ways related to the behavior of the counterexample, where {\tt LocFaults} reaches the condition $c_2$:
\begin{enumerate}
\item by following the branch induced by the condition $c_1$ ;
\item by following the opposite branch.
\end{enumerate}
The value of the label of arcs for case (1) (resp. (2)) is "\textit{next}" (resp. "\textit{devie}").
 
\begin{figure}[!ht]
\centering
\begin{tikzpicture}[->,>=stealth',shorten >=1pt,auto,node distance=1.2cm,thick,main node/.style={circle,draw,font=\sffamily\tiny}]
  \node[main node,draw=red!60, fill=red!5] (1) {1};
  \node[main node,draw=green!60, fill=green!5] (2) [below left of=1] {8};
  \node[main node,draw=blue!60, minimum size=6mm] (18) [below left of=1] {8};
  \node[main node,draw=green!60, fill=green!5] (3) [below left of=2] {9};
  \node[main node] (4) [below right of=3] {10};
  \node[main node,draw=green!60, fill=green!5] (6) [below right of=4] {11};
  \node[main node,draw=green!60, fill=green!5] (7) [below left of=6] {12};
  \node[main node,draw=red!60, fill=red!5] (9) [below right of=1] {2};
  \node[main node,draw=red!60, fill=red!5] (10) [below left of=9] {3};
  \node[main node,draw=red!60, fill=red!5] (11) [below right of=10] {4};
  \node[main node] (12) [below right of=11] {5};
  \node[main node] (13) [below left of=12] {6};
  \node[main node,draw=blue!60, fill=blue!5] (14) [below left of=3] {13};
  \node[main node,draw=blue!60, fill=blue!5] (15) [below right of=14] {14};
  \node[main node,draw=blue!60, fill=blue!5] (16) [below left of=15] {15};
  \node[main node,draw=blue!60, fill=blue!5] (17) [below right of=16] {16};  
  \node[main node,draw=red!60, fill=red!5] (8) [below right of=17] {7};
  \node[main node,draw=green!60, minimum size=6mm] (19) [below right of=17] {7};
  \node[main node,draw=red!60, fill=red!5] (20) [below right of=17] {7};
  \node[node distance=1.5cm,rectangle] (21) [below of=8] {{\scriptsize the path $<1,2,3,4,5,6,7,...,POST>$ is correct}};
  \node[node distance=0.4cm,rectangle] (22) [below of=21] {{\scriptsize the path $<1,8,9,10,11,12,7,...,POST>$ is correct}};
  \path[every node/.style={font=\sffamily\tiny}]
    (1) edge node [left] {next} (2)
    (18) edge node [left] {devie} (3) 
    (3) edge node [right] {devie} (4)
    (4) edge node [right] {next} (6)
    (6) edge node [left] {devie} (7)
    (7) edge node [left] {devie} (8)
    (1) edge node [right] {devie} (9)
    (9) edge node [left] {devie} (10)
    (10) edge node [right] {devie} (11)  
    (11) edge node [right] {devie} (12)
    (12) edge node [left] {next} (13)
    (13) edge node [left] {next} (8)         
    (3) edge node [left] {next} (14)
    (14) edge node [right] {devie} (15)
    (15) edge node [left] {devie} (16)
    (16) edge node [right] {devie} (17)
    (17) edge node [left] {devie} (8)     
    (8) edge node [left] {devie} (21);
\end{tikzpicture}
\label{cfgExempleAbstrait}
\caption{{\footnotesize Figure illustrating the execution of our algorithm on an example in which two minimal correction deviations are detected: $\{1,2,3,4,7\}$ and $\{8,9,11,12,7\}$, and one abandoned deviation: $\{8,13,14,15,16,7\}$. Knowing that the deviation of the condition "7" has corrected the program for the path $<1,2,3,4,5,6,7>$, and for the path $<1,8,9,10, 11,12,7>$. $POST$ in the figure is the postcondition.}}
\end{figure}
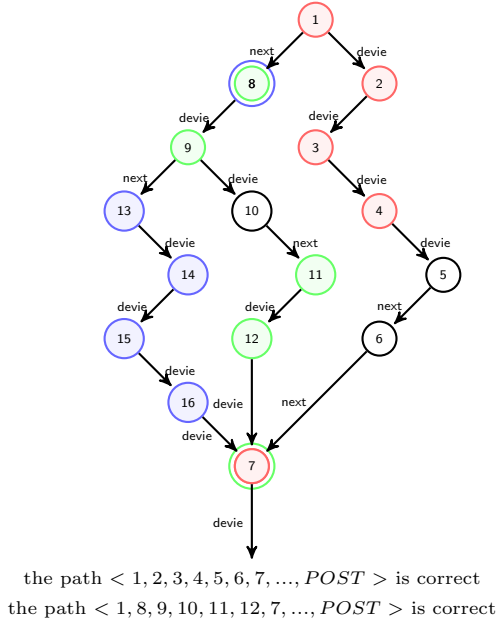

\begin{itemize}
\item At the step $k=5$, our algorithm has identified two MCDs of size equal to $5$:
\begin{enumerate}
\item $D_1=\{1,2,3,4,7\}$, the node "$7$" is marked by the value $5$ ;
\item $D_2=\{8,9,11,12,7\}$, it was allowed because the value of the marke of the node "$7$" is equal to the cardinality of $D_2$.  
\end{enumerate}

\item At the step $k=6$, the algorithm has suspended the following deviation $D_3=\{8,13,14,15,16,7\}$, because the cardinality of $D_3$ is strictly greater than the value of the label of the node "$7$". 
\end{itemize}

\section{Practical experience} 
To evaluate the scalability of our method, we compared its performance with that of {\tt BugAssist}\footnote{The tool BugAssist is available at : \url{http://bugassist.mpi-sws.org/}} on two sets benchmarks\footnote{The source code for all programs is available at : \url{http://www.i3s.unice.fr/~bekkouch/Benchs_Mohammed.html}}.
\begin{itemize}
\item[*] The first benchmark is illustrative, it contains a set of programs without loops;
\item[*]The second benchmark includes 19, 48 and 91 variations for respectively the programs BubbleSort, Sum and SquareRoot. These programs contain loops to study the scalability of our approach compared to {\tt BugAssist}. To increase the complexity of a program, we increase the number of iterations in loops in the execution of each tool; we use the same bound of unfolding loops for {\tt LocFaults} and {\tt BugAssist}.
\end{itemize}

To generate the CFG and the counterexample, we use the tool CPBPV~\cite{collavizza2010cpbpv} (Constraint-Programming Framework for Bounded Program Verification). {\tt LocFaults} and {\tt BugAssist} work respectively on Java and C programs. For a fair comparison, we built two equivalent versions for each program:
\begin{itemize}
\item[*] a version in Java annotated by a JML specification;
\item[*] a version in ANSI-C annotated by the same specification but in ACSL.
\end{itemize}
Both versions have the same numbers of lines of instructions, including errors. The precondition specifies the counterexample used for the program.

To calculate the MCSs, we used IBM ILOG MIP\footnote{IBM ILOG MIP is available at http://www-01.ibm.com/software/commerce/optimization/cplex-optimizer/} and CP\footnote{IBM ILOG CP OPTIMIZER is available at http://www-01.ibm.com/software/commerce/optimization/cplex-cp-optimizer/} solvers of CPLEX. We adapted and implemented the algorithm of Liffiton and Sakallah~\cite{liffiton2008algorithms}, see alg.~\ref{liffiton2008}. This implementation takes as input the infeasible set of constraints corresponding to the identified path ($C$), and $b_{mcs}$: the bound on the size of calculated MCSs. Each constraint $c_i$ in the system built $C$ is augmented by an indicator $y_i$  for giving $y_i  \rightarrow c_i $ in the new system of constraints $C '$. Assign to $y_i$ the value $True$ implies the constraint $c_i$; however, assign to $y_i$ value  $False$ implies the removal of the constraint $c_i$. A MCS is obtained by seeking an assignment that satisfies the constraint system with a minimal set of constraints indicators affected with $False$. To limit the number of constraints indicators that can be assigned with $False$, we use the constraint $AtMost({\neg y_1,\neg y_2,...,\neg y_n},k)$ (see the line $5$), the created system is noted in the algorithm $C'_{k}$ (line $5$). Each iteration of the \textsc{While}-loop (lines $6-19$) is allowed to find all MCSs of size $k$, $k$ is incremented by 1 after each iteration. After finding each MCS (lines $8-13$), a blocking constraint is added to $C'_k$ and $C'$ to prevent finding this new MCS in the next iterations (lines $15-16$). The first loop (lines $4-19$) is iterated until all MCSs of $C$ are generated ($C'$ becomes infeasible); it can also stop if the MCSs of size smaller or equal to $b_{mcs}$ are obtained ($k > b_{mcs}$).

\begin{algorithm}[h]
\begin{scriptsize} 
\label{MCS}
{\tt Function MCS}($C$,$b_{mcs}$)\\
\KwData{$C$: Infeasible set of  constraints, $b_{mcs}$: Integer}
\KwResult{$MCS$: List of MCSs in $C$ of a cardinality less than $b_{mcs}$}
\Begin{
$C'$ $\leftarrow$ \textsc{AddYVars}($C$); 
$MCS$ $\leftarrow$ $\emptyset$; 
$k$ $\leftarrow$ $1$; \\ 
\While{\textsc{SAT}($C'$) $\land$ $k \leq MCS_b$}{
$C'_{k}$ $\leftarrow$ $C'$ $\land$ \textsc{AtMost}($\{\neg y_{1},\neg y_{2},...,\neg y_{n}\}$,$k$)\\
\While{SAT($C'_{k}$)}{
$newMCS$ $\leftarrow$ $\emptyset$\\
\ForAll{indicator $y_{i}$}{
\% $y_{i}$ indicator of the constraint $c_{i} \in C$, and $val(y_{i})$ is the value of $y_{i}$ in the solution calculated for $C'_{k}$.\\
\Si{$val(y_{i})=0$}{
$newMCS$ $\leftarrow$ $newMCS$ $\cup$ $\{c_{i}\}$.\\
}
}
$MCS.add(newMCS)$.\\ 
$C'_{k}$ $\leftarrow$ $C'_{k}$ $\land$ \textsc{BlockingClause}($newMCS$) 
\\
$C'$ $\leftarrow$ $C'$ $\land$ \textsc{BlockingClause}($newMCS$) 
}
$k$ $\leftarrow$ $k$ + $1$\\
}
\Return $MCS$
}
\end{scriptsize}
\label{liffiton2008}
\caption{The algorithm of Liffiton and Sakallah}
\end{algorithm}

{\tt BugAssist} uses the tool CBMC~\cite{clarke2004tool} to generate the faulty trace and input data. For Max-SAT solver, we used MSUnCore2~\cite{marques2009msuncore}.

The experiments were performed with a processor Intel Core i7-3720QM 2.60 GHz with 8 GO of RAM.

\subsection{Benchmark without loops}

This part serves to illustrate the improvement in LocFaults to reduce the number of subsets of suspects instructions provided to the user: at a given step of the algorithm, the node in the CFG of the program that allows detect a MCD will be marked by the cardinality of the latter; in the next steps, the algorithm will not allow scanning an adjacency list of this node.

Our results\footnote{The table that shows the calculated MCSs by {\tt LocFaults} for the programs without loops are available at \url{http://www.i3s.unice.fr/~bekkouch/Benchs_Mohammed.html#rsb}} show that {\tt LocFaults} misses errors only for TritypeKO6. While {\tt BugAssist} misses errors for AbsMinusKO2, AbsMinusKO3, AbsMinusV2KO2 , TritypeKO , TriPerimetreKO, TriMultPerimetreKO and one of two errors in TritypeKO5. The times~\footnote{The tables that give the times of LocFaults and BugAssist for the programs without loops are available at \url{http://www.i3s.unice.fr/~bekkouch/
Benchs_Mohammed.html#rsba}.} of our tool are better compared to {\tt BugAssist} for programs with numerical calculation; they are close for the rest of programs.

We randomly take three programs as examples. And we consider the implementation of two versions of our algorithm with and without marking nodes named respectively {\tt LocFaultsV1} and {\tt LocFaultsV2}.

\begin{itemize}
\item Tables~\ref{MCSs1} and~\ref{time1} show respectively the suspects sets and times of {\tt LocFaultsV1} ;
\item Tables~\ref{MCSs2} and~\ref{time2} show respectively the suspects sets and times of {\tt LocFaultsV2}.
\end{itemize}
In tables~\ref{MCSs1} and~\ref{MCSs2}, we display the list of calculated MCSs and MCDs. The line number corresponding to the condition is underlined. Tables~\ref{time1} and~\ref{time2} give calculation times: $P$ is the pretreatment time which includes the translation of Java program into an abstract syntax tree with JDT tool (Eclipse Java devlopment tools), as well as the construction of CFG; $L$ is the time of the exploration of CFG and calculation of MCSs.     

{\tt LocFaultsV2} has significantly reduced the deviations generated and the time summing exploration of the CFG and calculation of MCSs by {\tt LocFaultsV1}, without losing the error; the localizations provided by {\tt LocFaultsV2} are more relevant. The eliminated lines of the table~\ref{MCSs2} are colored blue in the table\ref{MCSs1}. The improved time are shown in bold in the table~\ref{time1}. For example, for the program TritypeKO2, at step $1$ of the algorithm, {\tt LocFaultsV2} marks the node of condition $26$, $35$ and $53$ (from the counterexample, the program becomes correct by deviating each of these three conditions). This allows, at step $2$, to cancel the following deviations: $\{\uline{26},\uline{29}\}$, $\{\uline{26},\uline{35}\}$, $\{\uline{29},\uline{35}\}$, $\{\uline{32},\uline{35}\}$. Always in step $2$, {\tt LocFaultsV2} detects two minimal correction deviations more: $\{\uline{29},\uline{57}\}$, $\{\uline{32},\uline{44}\}$, the nodes $57$ and $44$ will be marked (the value of the mark is  $2$). At step $3$, no deviation is selected; for example, $\{\uline{29},\uline{32},\uline{44}\}$ is not considered because its cardinal is strictly superior to the mark value of the node $44$.

\begin{table*}
\begin{center}
\begin{tiny}
\begin{tabular}{|c|c|c|c|c|c|c|}
\hline
\multirow{2}{*}{Program} & \multirow{2}{*}{Counterexample} & \multirow{2}{*}{Errors} & \multicolumn{4}{|c|}{LocFaults} \\ 
\cline{4-7}
  &   &   & $= 0$ & $\leq 1$ & $\leq 2$ & $\leq 3$  \\   
\hline
\multirow{15}{*}{TritypeKO2}  &  \multirow{15}{*}{$\{i=2,j=2,k=4\}$}  &  \multirow{15}{*}{$53$}  &  \multirow{15}{*}{$\{54\}$}  &  $\{54\}$   & $\{54\}$   &  $\{54\}$   \\ 
\cline{6-7}
\cline{5-5}
&    &    &     & $\{\uline{21}\}$   & $\{\uline{21}\}$  &  $\{\uline{21}\}$   \\
\cline{6-7}
\cline{5-5}
&    &    &     & $\{\uline{26}\}$ & $\{\uline{26}\}$  &  $\{\uline{26}\}$  \\
\cline{5-7}
&    &    &     &  $\{\uline{35}\}$,$\{27\}$,$\{25\}$  & $\{\uline{35}\}$,$\{27\}$,$\{25\}$    & $\{\uline{35}\}$,$\{27\}$,$\{25\}$  \\
\cline{6-7}
\cline{5-5}
&    &    &     &  \multirow{11}{*}{$\{\uline{\textcolor{red}{53}}\}$,$\{25\}$,$\{27\}$}  &  $\{\uline{\textcolor{red}{53}}\}$,$\{25\}$,$\{27\}$     & $\{\uline{\textcolor{red}{53}}\}$,$\{25\}$,$\{27\}$       \\
\cline{6-7}
&    &    &     &     & \color{blue!90}$\{\uline{26},\uline{29}\}$  & \color{blue!90}$\{\uline{26},\uline{29}\}$     \\
\cline{6-7}
&    &    &     &     & \color{blue!90}$\{\uline{26},\uline{35}\}$,$\{25\}$ & \color{blue!90}$\{\uline{26},\uline{35}\}$,$\{25\}$    \\
\cline{6-7}
&    &    &     &     & \color{blue!90}$\{\uline{29},\uline{35}\}$,$\{30\}$,$\{25\}$,$\{27\}$  & \color{blue!90}$\{\uline{29},\uline{35}\}$,$\{30\}$,$\{25\}$,$\{27\}$     \\
\cline{6-7}
&    &    &     &     & $\{\uline{29},\uline{57}\}$,$\{30\}$,$\{27\}$,$\{25\}$   & $\{\uline{29},\uline{57}\}$,$\{30\}$,$\{27\}$,$\{25\}$   \\ 
\cline{6-7}
&    &    &     &     & \color{blue!90}$\{\uline{32},\uline{35}\}$,$\{33\}$,$\{25\}$,$\{27\}$   & \color{blue!90}$\{\uline{32},\uline{35}\}$,$\{33\}$,$\{25\}$,$\{27\}$   \\ 
\cline{6-7}
&    &    &     &     &  \multirow{6}{*}{$\{\uline{32},\uline{44}\}$,$\{33\}$,$\{25\}$,$\{27\}$}   &  $\{\uline{32},\uline{44}\}$,$\{33\}$,$\{25\}$,$\{27\}$   \\    
\cline{7-7}
&    &    &     &     &   & \color{blue!90}$\{\uline{26},\uline{29},\uline{35}\}$,$\{30\}$,$\{25\}$ \\ 
\cline{7-7}
&    &    &     &     &   & \color{blue!90}$\{\uline{26},\uline{32},\uline{35}\}$,$\{33\}$,$\{25\}$ \\
\cline{7-7}
&    &    &     &     &   & \color{blue!90}$\{\uline{26},\uline{32},\uline{57}\}$,$\{25\}$,$\{33\}$ \\
\cline{7-7}
\hhline{~~~~~~|6-6}
&    &    &     &     &   & \color{blue!90}$\{\uline{29},\uline{32},\uline{35}\}$,$\{33\}$,$\{25\}$,$\{27\}$,$\{30\}$ \\
\cline{7-7}
&    &    &     &     &   &  \color{blue!90}$\{\uline{29},\uline{32},\uline{44}\}$,$\{33\}$,$\{25\}$,$\{27\}$,$\{30\}$  \\
\hline
\multirow{14}{*}{TritypeKO4}  &  \multirow{14}{*}{$\{i=2,j=3,k=3\}$}  &  \multirow{14}{*}{$45$}  &  \multirow{14}{*}{$\{46\}$}  &  $\{46\}$   & $\{46\}$   &  $\{46\}$     \\
\cline{5-7}
&    &    &     &   \multirow{13}{*}{$\{\uline{\textcolor{red}{45}}\}$,$\{33\}$,$\{25\}$}  & $\{\uline{\textcolor{red}{45}}\}$,$\{33\}$,$\{25\}$   & $\{\uline{\textcolor{red}{45}}\}$,$\{33\}$,$\{25\}$  \\
\cline{6-7}
&    &    &     &     & $\{\uline{26},\uline{32}\}$   &  $\{\uline{26},\uline{32}\}$  \\
\cline{6-7}
&    &    &     &     & $\{\uline{29},\uline{32}\}$  & $\{\uline{29},\uline{32}\}$   \\
\cline{6-7}
&    &    &     &     &  \color{blue!90} $\{\uline{45},\uline{49}\}$,$\{33\}$,$\{25\}$  & \color{blue!90} $\{\uline{45},\uline{49}\}$,$\{33\}$,$\{25\}$   \\
\cline{6-7}
&    &    &     &     &  \color{blue!90}  & \color{blue!90} $\{\uline{45},\uline{53}\}$,$\{33\}$,$\{25\}$   \\
\cline{7-7}
&    &    &     &     &  \color{blue!90}  & \color{blue!90} $\{\uline{26},\uline{45},\uline{49}\}$,$\{33\}$,$\{25\}$,$\{27\}$   \\
\cline{7-7}
&    &    &     &     &  \color{blue!90}  & \color{blue!90} $\{\uline{26},\uline{45},\uline{53}\}$,$\{33\}$,$\{25\}$,$\{27\}$   \\
\cline{7-7}
&    &    &     &     &  \color{blue!90}  & \color{blue!90} $\{\uline{26},\uline{45},\uline{57}\}$,$\{33\}$,$\{25\}$,$\{27\}$   \\
\cline{7-7}
\hhline{~~~~~~|6-6}
&    &    &     &     &  \color{blue!90}  & \color{blue!90} $\{\uline{29},\uline{32},\uline{49}\}$,$\{30\}$,$\{25\}$   \\
\cline{7-7}
\hhline{~~~~~~|6-6}
&    &    &     &     &  \color{blue!90} $\{\uline{45},\uline{53}\}$,$\{33\}$,$\{25\}$ &  \color{blue!90} $\{\uline{29},\uline{45},\uline{49}\}$,$\{33\}$,$\{25\}$,$\{30\}$   \\
\cline{7-7}
\hhline{~~~~~~|6-6}
&    &    &     &     &  \color{blue!90}  & \color{blue!90} $\{\uline{29},\uline{45},\uline{53}\}$,$\{33\}$,$\{25\}$,$\{30\}$   \\
\cline{7-7}
\hhline{~~~~~~|6-6}
&    &    &     &     &  \color{blue!90}  & \color{blue!90} $\{\uline{29},\uline{45},\uline{57}\}$,$\{33\}$,$\{25\}$,$\{30\}$   \\
\cline{7-7}
&    &    &     &     &  \color{blue!90}  & $\{\uline{32},\uline{35},\uline{49}\}$,$\{25\}$    \\
\cline{7-7}
&    &    &     &     &  \color{blue!90}  & $\{\uline{32},\uline{35},\uline{53}\}$,$\{25\}$      \\
\cline{7-7}
&    &    &     &     &  \color{blue!90}  & $\{\uline{32},\uline{35},\uline{57}\}$,$\{25\}$     \\
\hline
\multirow{14}{*}{TriPerimetreKO3}  &  \multirow{14}{*}{$\{i=2,j=1,k=2\}$}  &  \multirow{14}{*}{$57$}   &  \multirow{14}{*}{$\{58\}$}  &  $\{58\}$  & $\{58\}$  &  $\{58\}$   \\
\cline{5-7}
&    &    &     & $\{\uline{22}\}$  &  $\{\uline{22}\}$   &  $\{\uline{22}\}$    \\
\cline{5-7}
&    &    &     &  $\{\uline{31}\}$  & $\{\uline{31}\}$  &  $\{\uline{31}\}$   \\
\cline{5-7}
&    &    &     &  $\{\uline{37}\}$,$\{32\}$,$\{27\}$   &  $\{\uline{37}\}$,$\{32\}$,$\{27\}$   &   $\{\uline{37}\}$,$\{32\}$,$\{27\}$  \\
\cline{5-7}
&    &    &     &  \multirow{10}{*}{$\{\textcolor{red}{\uline{57}}\}$,$\{32\}$,$\{27\}$}  &  $\{\textcolor{red}{\uline{57}}\}$,$\{32\}$,$\{27\}$  & $\{\textcolor{red}{\uline{57}}\}$,$\{32\}$,$\{27\}$   \\
\cline{6-7}
&    &    &     &     & \color{blue!90} $\{\uline{28},\uline{37}\}$,$\{32\}$,$\{27\}$,$\{29\}$   &  \color{blue!90} $\{\uline{28},\uline{37}\}$,$\{32\}$,$\{27\}$,$\{29\}$ \\ 
\cline{6-7}
&    &    &     &     & $\{\uline{28},\uline{61}\}$,$\{32\}$,$\{27\}$,$\{29\}$   &  $\{\uline{28},\uline{61}\}$,$\{32\}$,$\{27\}$,$\{29\}$    \\  
\cline{6-7}
&    &    &     &     & \color{blue!90} $\{\uline{31},\uline{37}\}$,$\{27\}$   & \color{blue!90} $\{\uline{31},\uline{37}\}$,$\{27\}$ \\
\cline{6-7}
&    &    &     &     & \color{blue!90} $\{\uline{34},\uline{37}\}$,$\{35\}$,$\{27\}$,$\{32\}$   & \color{blue!90} $\{\uline{34},\uline{37}\}$,$\{35\}$,$\{27\}$,$\{32\}$   \\
\cline{6-7}
&    &    &     &     & \multirow{7}{*}{$\{\uline{34},\uline{48}\}$,$\{35\}$,$\{32\}$,$\{27\}$}   &  $\{\uline{34},\uline{48}\}$,$\{35\}$,$\{32\}$,$\{27\}$   \\
\cline{7-7}
&    &    &     &     &    & \color{blue!90} $\{\uline{28},\uline{31},\uline{37}\}$,$\{29\}$,$\{27\}$  \\
\cline{7-7}
&    &    &     &     &    & \color{blue!90} $\{\uline{28},\uline{31},\uline{52}\}$,$\{29\}$,$\{27\}$ \\
\cline{7-7}
&    &    &     &     &    & \color{blue!90} $\{\uline{28},\uline{34},\uline{37}\}$,$\{35\}$,$\{27\}$,$\{29\}$,$\{32\}$ \\
\cline{7-7}
&    &    &     &     &    & \color{blue!90} $\{\uline{28},\uline{34},\uline{48}\}$,$\{35\}$,$\{27\}$,$\{29\}$,$\{32\}$ \\
\cline{7-7}
&    &    &     &     &    & \color{blue!90} $\{\uline{31},\uline{34},\uline{37}\}$,$\{27\}$,$\{35\}$ \\
\cline{7-7}
&    &    &     &     &    & \color{blue!90} $\{\uline{31},\uline{34},\uline{61}\}$,$\{27\}$,$\{35\}$ \\
\hline
\end{tabular}
\end{tiny}
\end{center}
\vspace{-0.5cm}
\caption{{\footnotesize MCSs and deviations identified by {\tt LocFaults} for programs without loops, without marking of nodes in the CFG}}
\vspace{-0.7cm}

\label{MCSs1}
\end{table*}

\begin{table}[!h]
\begin{center}
\begin{scriptsize}
\begin{tabular}{|c|c|c|c|c|c|}
\hline
\multirow{3}{*}{Program} & \multicolumn{5}{|c|}{LocFaults}  \\
\cline{2-6} &  \multirow{2}{*}{P}  &  \multicolumn{4}{|c|}{L}  \\
\cline{3-6}  &  & $= 0$ & $\leq 1$ & $\leq 2$ & $\leq 3$     \\
\hline
TritypeKO2  & $0,471$ & $0,023$  & $0,241$ & $2,529$ & $5,879$ \\
\hline
TritypeKO4 & $0,476$ & $0,022$ & $0,114$ & $0,348$ & $5,55$  \\
\hline
TriPerimetreKO3 & $0,487$  & $0,052$ &  $0,237$  & $2,468$ & $6,103$  \\
\hline
\end{tabular}
\end{scriptsize}
\end{center}
\vspace{-0.5cm}
\caption{{\footnotesize Computation time, for the results without marking of nodes in the CFG}}
\label{time1}
\end{table}

\begin{table*}
\begin{center}
\begin{tiny}
\begin{tabular}{|c|c|c|c|c|c|c|}
\hline
\multirow{2}{*}{Program} & \multirow{2}{*}{Counterexample} & \multirow{2}{*}{Errors} & \multicolumn{4}{|c|}{LocFaults} \\ 
\cline{4-7}
  &   &   & $= 0$ & $\leq 1$ & $\leq 2$ & $\leq 3$  \\
\hline
\multirow{7}{*}{TritypeKO2}  &  \multirow{7}{*}{$\{i=2,j=2,k=4\}$}  &  \multirow{7}{*}{$53$}  &  \multirow{7}{*}{$\{54\}$}  &   $\{54\}$  & $\{54\}$    &  $\{54\}$   \\ 
\cline{6-7}
\cline{5-5}
&    &    &     & $\{\uline{21}\}$    & $\{\uline{21}\}$   &  $\{\uline{21}\}$   \\
\cline{6-7}
\cline{5-5}
&    &    &     & $\{\uline{26}\}$  & $\{\uline{26}\}$   &  $\{\uline{26}\}$  \\
\cline{5-7}
&    &    &     &  $\{\uline{35}\}$,$\{27\}$,$\{25\}$   & $\{\uline{35}\}$,$\{27\}$,$\{25\}$   & $\{\uline{35}\}$,$\{27\}$,$\{25\}$  \\
\cline{6-7}
\cline{5-5}
&    &    &     & \multirow{3}{*}{$\{\uline{\textcolor{red}{53}}\}$,$\{25\}$,$\{27\}$}  &   $\{\uline{\textcolor{red}{53}}\}$,$\{25\}$,$\{27\}$  & $\{\uline{\textcolor{red}{53}}\}$,$\{25\}$,$\{27\}$       \\
\cline{6-7}
&    &    &     &     & $\{\uline{29},\uline{57}\}$,$\{30\}$,$\{27\}$,$\{25\}$   &  $\{\uline{29},\uline{57}\}$,$\{30\}$,$\{27\}$,$\{25\}$    \\
\cline{6-7}
&    &    &     &     &  $\{\uline{32},\uline{44}\}$,$\{33\}$,$\{25\}$, $\{27\}$  &  $\{\uline{32},\uline{44}\}$,$\{33\}$,$\{25\}$, $\{27\}$    \\   
\hline
\multirow{7}{*}{TritypeKO4}  &  \multirow{7}{*}{$\{i=2,j=3,k=3\}$}  &  \multirow{7}{*}{$45$}  &  \multirow{7}{*}{$\{46\}$}  &  $\{46\}$   & $\{46\}$   &  $\{46\}$    \\
\cline{5-7}
&    &    &     &   \multirow{6}{*}{$\{\uline{\textcolor{red}{45}}\}$,$\{33\}$,$\{25\}$}  & $\{\uline{\textcolor{red}{45}}\}$,$\{33\}$,$\{25\}$   & $\{\uline{\textcolor{red}{45}}\}$,$\{33\}$,$\{25\}$  \\
\cline{6-7}
&    &    &     &     & $\{\uline{26},\uline{32}\}$   &  $\{\uline{26},\uline{32}\}$  \\
\cline{6-7}
&    &    &     &     &  \multirow{4}{*}{$\{\uline{29},\uline{32}\}$}  & $\{\uline{29},\uline{32}\}$   \\
\cline{7-7}
&    &    &     &     &    & $\{\uline{32},\uline{35},\uline{49}\}$,$\{25\}$    \\
\cline{7-7}
&    &    &     &     &    & $\{\uline{32},\uline{35},\uline{53}\}$,$\{25\}$      \\
\cline{7-7}
&    &    &     &     &    & $\{\uline{32},\uline{35},\uline{57}\}$,$\{25\}$     \\
\hline
\multirow{7}{*}{TriPerimetreKO3}  &  \multirow{7}{*}{$\{i=2,j=1,k=2\}$}  &  \multirow{7}{*}{$57$}   &  \multirow{7}{*}{$\{58\}$}  &  $\{58\}$  & $\{58\}$  &  $\{58\}$   \\
\cline{5-7}
&    &    &     & $\{\uline{22}\}$  &  $\{\uline{22}\}$   &  $\{\uline{22}\}$    \\
\cline{5-7}
&    &    &     &  $\{\uline{31}\}$  & $\{\uline{31}\}$  &  $\{\uline{31}\}$   \\
\cline{5-7}
&    &    &     &  $\{\uline{37}\}$,$\{32\}$,$\{27\}$   &  $\{\uline{37}\}$,$\{32\}$,$\{27\}$   &   $\{\uline{37}\}$,$\{32\}$,$\{27\}$  \\
\cline{5-7}
&    &    &     &  \multirow{3}{*}{$\{\textcolor{red}{\uline{57}}\}$,$\{32\}$,$\{27\}$}  &  $\{\textcolor{red}{\uline{57}}\}$,$\{32\}$,$\{27\}$  & $\{\textcolor{red}{\uline{57}}\}$,$\{32\}$,$\{27\}$   \\
\cline{6-7}
&    &    &     &     & $\{\uline{28},\uline{61}\}$,$\{32\}$,$\{27\}$,$\{29\}$   &  $\{\uline{28},\uline{61}\}$,$\{32\}$,$\{27\}$,$\{29\}$    \\  
\cline{6-7}
&    &    &     &     & $\{\uline{34},\uline{48}\}$,$\{35\}$,$\{32\}$,$\{27\}$   &  $\{\uline{34},\uline{48}\}$,$\{35\}$,$\{32\}$,$\{27\}$   \\
\hline
\end{tabular}
\end{tiny}
\end{center}
\vspace{-0.5cm}
\caption{{\footnotesize MCSs and MCDs identified by {\tt LocFaults} for programs without loops, with marking of nodes in the CFG}}
\vspace{-0.7cm}
\label{MCSs2}
\end{table*}

\begin{table}[!h]
\begin{center}
\begin{scriptsize}
\begin{tabular}{|c|c|c|c|c|c|}
\hline
\multirow{3}{*}{Program} & \multicolumn{5}{|c|}{LocFaults}  \\
\cline{2-6} &  \multirow{2}{*}{P}  &  \multicolumn{4}{|c|}{L}  \\
\cline{3-6}  &  & $= 0$ & $\leq 1$ & $\leq 2$ & $\leq 3$    \\
\hline
TritypeKO2  & $0,496$ & $0,022$  & $0,264$ & $\textbf{1,208}$ & $\textbf{1,119}$  \\
\hline
TritypeKO4 & $0,481$ & $0,021$ & $0,106$ & $\textbf{0,145}$ & $\textbf{1,646}$  \\
\hline
TriPerimetreKO3 & $0,485$  & $0,04$ &  $0,255$  & $\textbf{1,339}$ & $\textbf{1,219}$  \\
\hline
\end{tabular}
\end{scriptsize}
\end{center}
\vspace{-0.5cm}
\caption{{\footnotesize Computation time, for the results with marking of nodes in the CFG}}
\label{time2}
\end{table}

\subsection{Benchmarks with loops}
These benchmarks are used to measure the scalability of {\tt LocFaults} compared to {\tt BugAssist} for programs with loops, depending on the increase of unfolding $b$. We took three programs with loops : BubbleSort, Sum, and SquareRoot. We have caused the \textit{Off-by-one} bug in each of them. The benchmark for each program is created by increasing the number of unfolding $b$. $b$ is equal to the number of iterations through the loop in the worst case. We also vary the number of deviated conditions for {\tt LocFaults} from $0$ to $3$.

We used the MIP solver of CPLEX for BubbleSort. For Sum and SquareRoot, we collaborate the two solvers of CPLEX (CP and MIP) during the localization process. Indeed, during the collection of constraints, we use a variable to keep the information on the type of building CSP. When {\tt LocFaults} detects an erroneous path\footnote{An erroneous path is the one on which we identify MCSs.} and prior to the calculation of MCSs, it takes the good solver depending on the type of CSP corresponding to this path : if it is non-linear, it uses the CP OPTIMIZER solver; otherwise it uses the MIP solver.

For each benchmark, we presented an extract of the table containing the computation time\footnote{Full tables are available at \url{http://www.i3s.unice.fr/~bekkouch/Benchs_Mohammed.html#ravb}, the sources of these results are available at \url{http://www.i3s.unice.fr/~bekkouch/Benchs_Mohammed.html#sr}} (columns $P$ and $L$ show respectively the time of pretreatment and calculating of MCSs), and the graph which corresponds to the time of calculation of MCSs.
 
\subsubsection{BubbleSort benchmark}
BubbleSort is an implementation of the bubble sort algorithm. This program contains two nested loops; its average complexity is $O(n^2)$, where $n$ is the size of the table sorted : the bubble sort is considered among the worst sort algorithms. The erroneous statement in the program causes the program to sort input array by considering only its $n-1$ first elements. The malfunction of BubbleSort is due to the insufficient number of iterations performed by the loop. This is due to the faulty initialization of the variable i :  i = tab.length - 1; the instruction should be i = tab.length. 
\begin{table}[!h]
{\fontsize{1pt}{1pt}\selectfont
\tabcolsep=2pt
\begin{center}
\begin{scriptsize}
\begin{tabular}{|c|c|c|c|c|c|c|c|c|}
\hline
\multirow{3}{*}{Programs} & \multirow{3}{*}{b} & \multicolumn{5}{|c|}{LocFaults} & \multicolumn{2}{c|}{BugAssist} \\
\cline{3-9} & & \multirow{2}{*}{P}  &  \multicolumn{4}{c|}{L} & \multirow{2}{*}{P} & \multirow{2}{*}{L} \\
\cline{4-7}  & & & $= 0$ & $\leq 1$ & $\leq 2$ & $\leq 3$ &   &    \\
\hline
V0 & $4$ & $ 0.751$  & $ 0.681$  & $ 0.56$  & $ 0.52$  & $ 0.948$  & $0.34$  & $55.27$ \\
\hline
V1 & $5$ & $ 0.813$  & $ 0.889$  & $ 0.713$  & $ 0.776$  & $ 1.331$  & $0.22$  & $125.40$ \\
\hline
V2 & $6$ & $ 1.068$  & $ 1.575$  & $ 1.483$  & $ 1.805$  & $ 4.118$  & $0.41$  & $277.14$ \\
\hline
V3 & $7$ & $ 1.153$  & $ 0.904$  & $ 0.85$  & $ 1.597$  & $ 12.67$  & $0.53$  & $612.79$ \\
\hline
V4 & $8$ & $ 0.842$  & $ 6.509$  & $ 6.576$  & $ 8.799$  & $ 116.347$  & $1.17$  & $1074.67$ \\
\hline
V5 & $9$ & $ 1.457$  & $ 18.797$  & $ 18.891$  & $ 21.079$  & $ 492.178$  & $1.24$  & $1665.62$ \\
\hline
V6 & $10$ & $ 0.941$  & $ 28.745$  & $ 29.14$  & $ 35.283$  & $ 2078.445$  & $1.53$  & $2754.68$ \\
\hline
V7 & $11$ & $ 0.918$  & $ 59.894$  & $ 65.289$  & $ 74.93$  & $ 4916.434$  & $3.94$  & $7662.90$ \\
\hline

\end{tabular}
\end{scriptsize}
\end{center}
\vspace{-0.5cm}
\caption{Computation time for benchmark BubbleSort}
\label{TimesLB}
}
\end{table}

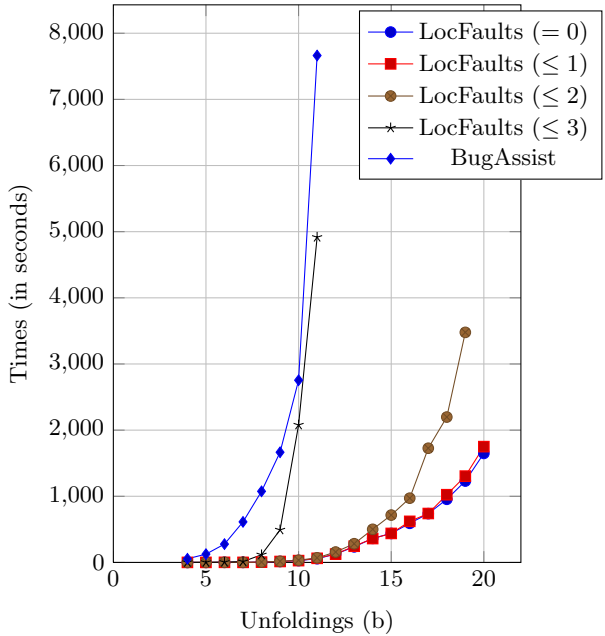
\begin{figure}[!h]
\begin{tikzpicture} 

\begin{axis}
[ 
  xlabel=Unfoldings (b), ylabel=Times (in seconds),ymin=0,xmin=0,
  height=9cm, width=7cm, grid=major,legend style={at={(1.20,0.99)}},
] 
\addplot coordinates { 
 
(4, 0.561)
(5, 0.597)
(6, 1.461)
(7, 0.813)
(8, 4.787)
(9, 14.234)
(10, 27.389)
(11, 56.008)
(12, 126.439)
(13, 235.282)
(14, 363.627)
(15, 437.994)
(16, 591.28)
(17, 737.541)
(18, 954.475)
(19, 1230.099)
(20, 1647.91)
};
\addlegendentry{LocFaults ($= 0$)} 

\addplot coordinates { 
  
(4, 0.553)
(5, 0.627)
(6, 1.496)
(7, 0.852)
(8, 4.911)
(9, 14.228)
(10, 27.608)
(11, 62.198)
(12, 126.233)
(13, 244.805)
(14, 360.651)
(15, 438.549)
(16, 621.072)
(17, 739.541)
(18, 1023.731)
(19, 1305.219)
(20, 1750.644)
}; 
\addlegendentry{LocFaults ($\leq 1$)} 

\addplot coordinates { 
  
(4, 0.508)
(5, 0.762)
(6, 1.75)
(7, 1.468)
(8, 6.01)
(9, 16.753)
(10, 33.573)
(11, 69.591)
(12, 157.238)
(13, 282.796)
(14, 500.626)
(15, 715.594)
(16, 971.357)
(17, 1726.373)
(18, 2197.53)
(19, 3477.862)
}; 
\addlegendentry{LocFaults ($\leq 2$)}

\addplot coordinates { 
  
(4, 0.948)
(5, 1.331)
(6, 4.118)
(7, 12.67)
(8, 116.347)
(9, 492.178)
(10, 2078.445)
(11, 4916.434)
}; 
\addlegendentry{LocFaults ($\leq 3$)}

\addplot coordinates { 
  
(4,55.27)
(5,125.40)
(6,277.14)
(7,612.79)
(8,1074.67)
(9,1665.62)
(10,2754.68)
(11,7662.90)
};
\addlegendentry{BugAssist}
\end{axis} 
\end{tikzpicture}
\vspace{-0.3cm}
\caption{Comparison of the evolution of times of different versions of {\tt LocFaults} and of {\tt BugAssist} for the benchmark BubbleSort, by increasing the unwinding loop limit.}
\label{LFvsBA}
\end{figure}
The times of {\tt LocFaults} and {\tt BugAssist} for the benchmark BubbleSort are presented in the table~\ref{TimesLB}. The graph illustrates the increase in times of different versions of {\tt LocFaults} and of {\tt BugAssist} depending on the number of unfolding is given in Figure~\ref{LFvsBA}.

The runtime of {\tt LocFaults} and of {\tt BugAssist} grows exponentially with the number of unfoldings; the times of {\tt BugAssist} are always the greatest. We can consider that {\tt BugAssist} is ineffective for this benchmark. The different versions of {\tt LocFaults} (with at most $3$, $2$, $1$, and $0$ conditions deviated) remain usable up to a certain unfolding. The number of unfolding beyond which growth time of {\tt BugAssist} becomes redhibitory is lower than that of {\tt LocFaults}, that of {\tt LocFaults} with at most $3$ conditions diviated is lower than that of {\tt LocFaults} with at most $2$ conditions diviated which is also lower than that of {\tt LocFaults} with at most $1$ conditions diviated. The times of {\tt LocFaults} with at most $1$ and $0$ conditions diviated are almost the same.

\subsubsection{SquareRoot and Sum benchmarks}
The program SquareRoot (see fig.~\ref{SquareRoot}) permits to find the integer part of the square root of the integer 50. An error is injected at the line 13, which leads to return the value 8; while the program must return 7. This program has been used in the paper describing the approach {\tt BugAssist}, it contains a linear numerical calculation in its loop and nonlinear in its postcondition.

\begin{figure}[!h]
    \center
\begin{lstlisting}{}
class SquareRoot{
  /*@ ensures((res*res<=val) && (res+1)*(res+1)>val);*/
  int SquareRoot()
    {
        int val = 50;
        int i = 1;
        int v = 0;
        int res = 0;
        while (v < val){
             v = v + 2*i + 1;
             i= i + 1; 
        }
        res = i; /*error: the instruction should be res = i - 1*/         
        return res;
   }
}
\end{lstlisting}
\vspace{-0.9cm}
\caption{The program SquareRoot}
\label{SquareRoot}
\end{figure}

With an unwinding limit of $50$, {\tt BugAssist} calculates for this program the following suspicious instructions: $\{9, 10, 11$ $, 13\}$. The time of localization is $36,16s$ and the pretreatment time is $0,12s$.

{\tt LocFaults} displays a suspicious instruction by indicating both its location in the program (instruction line), the line of the condition and the iteration of each loop leading to this instruction. For example, $\{9:2.11\}$ corresponds to the instruction that is on line $11$ in the program, the latter is in a loop whose line of the stop condition is $9$ and the iteration number is $2$. The sets suspected by {\tt LocFaults} are provided in the table~\ref{McdMcsSquareRoot}.

\begin{table}
\begin{scriptsize}
\begin{tabular}{|c|c|}
  \hline
  \textit{MCDs} & \textit{MCSs}  \\
  \hline
  $\emptyset$ & $\{5\}$,$\{6\}$,$\{9:1.11\}$, $\{9:2.11\}$,$\{9:3.11\}$,  \\
   & $\{9:4.11\}$,$\{9:5.11\}$,$\{9:6.11\}$,$\{9:7.11\}$,$\{\textcolor{red}{13}\}$  \\
  \hline            
  \multirow{2}{*}{$\{9:7\}$} & $\{5\}$,$\{6\}$,$\{7\}$,$\{9:1.10\}$,$\{9:2.10\}$,$\{9:3.10\}$,  \\ 
  & $\{9:4.10\}$,$\{9:5.10\}$, $\{9:6.10\}$,$\{9:1.11\}$, \\
  & $\{9:2.11\}$,$\{9:3.11\}$,$\{9:4.11\}$,$\{9:5.11\}$, $\{9:6.11\}$ \\ 
  \hline
\end{tabular}
\end{scriptsize}
\caption{MCD and MCSs calculated by {\tt LocFaults} for SquareRoot.}
\label{McdMcsSquareRoot}
\end{table}

The pretreatment time is $0,769s$. The time during the exploration of the CFG and the calculation of MCSs is $1,299s$. We studied the times of {\tt LocFaults} and {\tt BugAssist} of values of $val$ ranging from $10$ to $100$ (the number of unfolding $b$ used is equal to $val$), to study the combinatorial behavior of each tool for this program.

\begin{table}[h]
{\fontsize{1pt}{1pt}\selectfont
\tabcolsep=2pt
\begin{center}
\begin{scriptsize}
\begin{tabular}{|c|c|c|c|c|c|c|c|c|}
\hline
\multirow{3}{*}{Programs} & \multirow{3}{*}{b} & \multicolumn{5}{|c|}{LocFaults} & \multicolumn{2}{c|}{BugAssist} \\
\cline{3-9} & & \multirow{2}{*}{P}  &  \multicolumn{4}{c|}{L} & \multirow{2}{*}{P} & \multirow{2}{*}{L} \\
\cline{4-7}  & & & $= 0$ & $\leq 1$ & $\leq 2$ & $\leq 3$ &   &    \\
\hline
V0 & $10$ & $ 1.096$  & $ 1.737 $  & $ 2.098 $  & $ 2.113 $  & $ 2.066 $  & $0.05$  & $3.51$ \\
\hline
V10 & $20$ & $ 0.724$  & $ 0.974 $  & $ 1.131 $  & $ 1.117 $  & $ 1.099 $  & $0.05$  & $6.54$ \\
\hline
V20 & $30$ & $ 0.771$  & $ 1.048 $  & $ 1.16 $  & $ 1.171 $  & $ 1.223 $  & $0.08$  & $12.32$ \\
\hline
V30 & $40$ & $ 0.765$  & $ 1.048 $  & $ 1.248 $  & $ 1.266 $  & $ 1.28 $  & $0.09$  & $23.35$ \\
\hline
V40 & $50$ & $ 0.769$  & $ 1.089 $  & $ 1.271 $  & $ 1.291 $  & $ 1.299 $  & $0.12$  & $36.16$ \\
\hline
V50 & $60$ & $ 0.741$  & $ 1.041 $  & $ 1.251 $  & $ 1.265 $  & $ 1.281 $  & $0.14$  & $38.22$ \\
\hline
V70 & $80$ & $ 0.769$  & $ 1.114 $  & $ 1.407 $  & $ 1.424 $  & $ 1.386 $  & $0.19$  & $57.09$ \\
\hline
V80 & $90$ & $ 0.744$  & $ 1.085 $  & $ 1.454 $  & $ 1.393 $  & $ 1.505 $  & $0.22$  & $64.94$ \\
\hline
V90 & $100$ & $ 0.791$  & $ 1.168 $  & $ 1.605 $  & $ 1.616 $  & $ 1.613 $  & $0.24$  & $80.81$ \\
\hline

\end{tabular}
\end{scriptsize}
\end{center}
\vspace{-0.3cm}
\caption{The computation time for the benchmark SquareRoot}
\label{TableLFvsBASquareRoot}
}
\end{table}

\begin{table}[!h]
{\fontsize{1pt}{1pt}\selectfont
\tabcolsep=2pt
\begin{center}
\begin{scriptsize}
\begin{tabular}{|c|c|c|c|c|c|c|c|c|}
\hline
\multirow{3}{*}{Programs} & \multirow{3}{*}{b} & \multicolumn{5}{|c|}{LocFaults} & \multicolumn{2}{c|}{BugAssist} \\
\cline{3-9} & & \multirow{2}{*}{P}  &  \multicolumn{4}{c|}{L} & \multirow{2}{*}{P} & \multirow{2}{*}{L} \\
\cline{4-7}  & & & $= 0$ & $\leq 1$ & $\leq 2$ & $\leq 3$ &   &    \\
\hline
V0 & $6$ & $ 0.765$  & $ 0.427$  & $ 0.766$  & $ 0.547$  & $ 0.608$  & $0.04$  & $2.19$ \\
\hline
V10 & $16$ & $ 0.9$  & $ 0.785$  & $ 1.731$  & $ 1.845$  & $ 1.615$  & $0.08$  & $17.88$ \\
\hline
V20 & $26$ & $ 1.11$  & $ 1.449$  & $ 7.27$  & $ 7.264$  & $ 6.34$  & $0.12$  & $53.85$ \\
\hline
V30 & $36$ & $ 1.255$  & $ 0.389$  & $ 8.727$  & $ 4.89$  & $ 4.103$  & $0.13$  & $108.31$ \\
\hline
V40 & $46$ & $ 1.052$  & $ 0.129$  & $ 5.258$  & $ 5.746$  & $ 13.558$  & $0.23$  & $206.77$ \\
\hline
V50 & $56$ & $ 1.06$  & $ 0.163$  & $ 7.328$  & $ 6.891$  & $ 6.781$  & $0.22$  & $341.41$ \\
\hline
V60 & $66$ & $ 1.588$  & $ 0.235$  & $ 13.998$  & $ 13.343$  & $ 14.698$  & $0.36$  & $593.82$ \\
\hline
V70 & $76$ & $ 0.82$  & $ 0.141$  & $ 10.066$  & $ 9.453$  & $ 10.531$  & $0.24$  & $455.76$ \\
\hline
V80 & $86$ & $ 0.789$  & $ 0.141$  & $ 13.03$  & $ 12.643$  & $ 12.843$  & $0.24$  & $548.83$ \\
\hline
V90 & $96$ & $ 0.803$  & $ 0.157$  & $ 34.994$  & $ 28.939$  & $ 18.141$  & $0.31$  & $785.64$ \\
\hline
\end{tabular}
\end{scriptsize}
\end{center}
\vspace{-0.3cm}
\caption{The computation time for the benchmark Sum}
\label{TableLFvsBASum}
}
\end{table}

\begin{figure}[!h]
\begin{tikzpicture} 

\begin{axis}
[ 
  xlabel=Unfoldings (b), ylabel=Times (in seconds),ymin=0,xmin=0,
  height=9cm, width=7cm, grid=major,legend style={at={(0.70,0.99)}},
] 

\addplot coordinates { 
  
(10, 2.066 )
(11, 2.15 )
(12, 2.074 )
(13, 2.384 )
(14, 1.881 )
(15, 1.871 )
(16, 2.386 )
(17, 2.024 )
(18, 1.149 )
(19, 1.128 )
(20, 1.099 )
(21, 1.124 )
(22, 1.124 )
(23, 1.154 )
(24, 1.132 )
(25, 1.199 )
(26, 1.22 )
(27, 1.212 )
(28, 1.147 )
(29, 1.282 )
(30, 1.223 )
(31, 1.19 )
(32, 1.166 )
(33, 1.155 )
(34, 1.16 )
(35, 1.194 )
(36, 1.226 )
(37, 1.222 )
(38, 1.243 )
(39, 1.249 )
(40, 1.28 )
(41, 1.247 )
(42, 1.257 )
(43, 1.269 )
(44, 1.286 )
(45, 1.251 )
(46, 1.199 )
(47, 1.277 )
(48, 1.226 )
(49, 1.265 )
(50, 1.299 )
(51, 1.31 )
(52, 1.286 )
(53, 1.217 )
(54, 1.291 )
(55, 1.336 )
(56, 1.281 )
(57, 1.218 )
(58, 1.275 )
(59, 1.307 )
(60, 1.281 )
(61, 1.301 )
(62, 1.321 )
(63, 1.306 )
(64, 1.368 )
(65, 1.328 )
(66, 1.386 )
(67, 1.423 )
(68, 1.417 )
(69, 1.292 )
(70, 1.408 )
(71, 1.378 )
(72, 1.31 )
(73, 1.404 )
(74, 1.401 )
(75, 1.297 )
(76, 1.443 )
(77, 1.361 )
(78, 1.361 )
(79, 1.391 )
(80, 1.386 )
(81, 1.545 )
(82, 1.293 )
(83, 1.427 )
(84, 1.376 )
(85, 1.373 )
(86, 1.487 )
(87, 1.576 )
(88, 1.494 )
(89, 1.518 )
(90, 1.505 )
(91, 1.374 )
(92, 1.361 )
(93, 1.397 )
(94, 1.41 )
(95, 1.422 )
(96, 1.437 )
(97, 1.425 )
(98, 1.488 )
(99, 1.456 )
(100, 1.613 )
}; 
\addlegendentry{LocFaults ($\leq 3$)}

\addplot coordinates { 
  
(10,3.51)
(11,3.35)
(12,3.74)
(13,4.88)
(14,5.30)
(15,5.55)
(16,6.60)
(17,6.50)
(18,6.08)
(19,5.64)
(20,6.54)
(21,6.66)
(22,6.71)
(23,7.70)
(24,7.67)
(25,12.58)
(26,10.76)
(27,12.78)
(28,12.74)
(29,13.64)
(30,12.32)
(31,15.74)
(32,16.45)
(33,17.83)
(34,17.68)
(35,20.99)
(36,19.90)
(37,19.72)
(38,22.55)
(39,22.69)
(40,23.35)
(41,22.86)
(42,23.64)
(43,27.51)
(44,27.19)
(45,29.28)
(46,30.27)
(47,29.90)
(48,30.00)
(49,36.32)
(50,36.16)
(51,34.46)
(52,34.09)
(53,42.99)
(54,39.28)
(55,38.81)
(56,39.42)
(57,42.27)
(58,42.87)
(59,44.93)
(60,38.22)
(61,46.18)
(62,44.53)
(63,45.45)
(64,47.91)
(65,51.55)
(66,50.51)
(67,51.83)
(68,53.82)
(69,56.22)
(70,55.15)
(71,56.52)
(72,56.00)
(73,55.55)
(74,57.12)
(75,59.67)
(76,53.98)
(77,57.38)
(78,56.68)
(79,57.75)
(80,57.09)
(81,69.00)
(82,64.40)
(83,67.80)
(84,67.25)
(85,65.55)
(86,69.78)
(87,68.32)
(88,68.66)
(89,67.41)
(90,64.94)
(91,70.76)
(92,66.88)
(93,70.20)
(94,66.38)
(95,66.82)
(96,65.36)
(97,70.84)
(98,72.02)
(99,71.42)
(100,80.81)
};
\addlegendentry{BugAssist}

\end{axis} 
\end{tikzpicture}
\vspace{-0.5cm}
\caption{Comparison of the evolution of times of {\tt LocFaults} with at most $3$ conditions deviated and of {\tt BugAssist} for the benchmark SquareRoot, by increasing the unwinding loop limit.}
\label{GrapheLFvsBASquareRoot}
\end{figure}

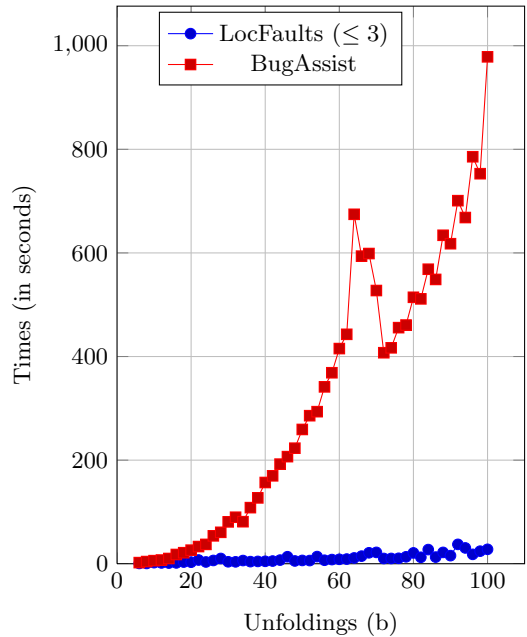
\begin{figure}[!h]
\begin{tikzpicture} 

\begin{axis}
[ 
  xlabel=Unfoldings (b), ylabel=Times (in seconds),ymin=0,xmin=0,
  height=9cm, width=7cm, grid=major,legend style={at={(0.70,0.99)}},
] 

\addplot coordinates { 
  
(6, 0.608)
(8, 0.583)
(10, 2.103)
(12, 1.567)
(14, 1.308)
(16, 1.615)
(18, 3.279)
(20, 3.0)
(22, 7.216)
(24, 3.436)
(26, 6.34)
(28, 10.17)
(30, 3.809)
(32, 3.681)
(34, 6.187)
(36, 4.103)
(38, 4.502)
(40, 4.706)
(42, 5.202)
(44, 7.047)
(46, 13.558)
(48, 5.396)
(50, 6.384)
(52, 6.163)
(54, 13.819)
(56, 6.781)
(58, 8.055)
(60, 8.553)
(62, 9.03)
(64, 11.205)
(66, 14.698)
(68, 21.123)
(70, 21.921)
(72, 10.175)
(74, 10.206)
(76, 10.531)
(78, 13.366)
(80, 20.915)
(82, 12.487)
(84, 27.336)
(86, 12.843)
(88, 21.9)
(90, 15.67)
(92, 37.395)
(94, 30.523)
(96, 18.141)
(98, 24.283)
(100, 27.856)
}; 
\addlegendentry{LocFaults ($\leq 3$)}

\addplot coordinates { 
  
(6,2.19)
(8,4.22)
(10,5.98)
(12,7.20)
(14,10.14)
(16,17.88)
(18,21.31)
(20,26.24)
(22,33.18)
(24,37.36)
(26,53.85)
(28,60.68)
(30,80.80)
(32,89.79)
(34,81.19)
(36,108.31)
(38,127.10)
(40,156.84)
(42,169.73)
(44,192.34)
(46,206.77)
(48,223.07)
(50,259.24)
(52,285.80)
(54,293.56)
(56,341.41)
(58,368.68)
(60,415.34)
(62,442.89)
(64,674.55)
(66,593.82)
(68,598.82)
(70,527.41)
(72,407.33)
(74,416.81)
(76,455.76)
(78,460.84)
(80,514.36)
(82,511.29)
(84,568.54)
(86,548.83)
(88,634.07)
(90,617.77)
(92,700.87)
(94,668.30)
(96,785.64)
(98,752.92)
(100,978.51)
};
\addlegendentry{BugAssist}

\end{axis} 
\end{tikzpicture}
\vspace{-0.4cm}
\caption{Comparison of the evolution of times of {\tt LocFaults} with at most $3$ conditions deviated and of {\tt BugAssist} for the benchmark Sum, by increasing the unwinding loop limit.}
\label{GrapheLFvsBASum}
\end{figure}

The program Sum takes a positive integer $n$ from the user, and it calculates the value of $\sum_{i=1}^{n}i$. The postcondition specifies that sum. The error in Sum is in the condition of its loop. It causes to calculate the sum $\sum_{i=1}^{n-1}i$ instead of $\sum_{i=1}^{n}i$. This program contains linear numerical instructions in the core of the loop, and a nonlinear postcondition.

The results in time for SquareRoot and Sum benchmarks are shown in the tables respectively~\ref{TableLFvsBASquareRoot} and~\ref{TableLFvsBASum}. We also designed the graph that corresponds to the result of each benchmark, see respectively the graphs in Figure~\ref{GrapheLFvsBASquareRoot} and~\ref{GrapheLFvsBASum}. The execution time of {\tt BugAssist} grows rapidly; the times of {\tt LocFaults} are almost constant. The times of {\tt LocFaults} with at most $0$, $1$, and $2$ conditions deviated are similar to those of {\tt LocFaults} with at most $3$ conditions deviated.

\section{Conclusion}
The method {\tt LocFaults} detects the suspicious subsets by analyzing the paths of the CFG to find the MCDs and MCSs from each MCD; it uses constraint solvers. The method {\tt BugAssit} calculates the merger of MCSs of the program  by transforming the whole program into a Boolean formula; it uses Max-SAT solvers. Both methods work by starting from a counterexample. In this paper, we presented an exploration of scalability of {\tt LocFaults}, particularly on the treatment of loops with the \textit{Off-by-one} bug. The first results show that {\tt LocFaults} is more effective than {\tt BugAssist} on programs with loops.
 The times of {\tt BugAssist} rapidly increase with the number of unfolding.
 
As part of our future work, we plan to validate our results on programs with more complex loops. We envisage to compare the performance of {\tt LocFaults} with existing statistical methods. To improve our tool, we develop an interactive version that provides the suspect subsets, one after the other : we want to take advantage of the user's knowledge to select the conditions that should be deviated. We also reflect on how to extend our method to treat numerical instructions with calculation on floating-point.  

\section{Acknowledgments}
Thanks to Bertrand Neveu for his careful reading and helpful comments on this paper. Thanks to Michel Rueher and H\'el\`ene Collavizza for their interesting remarks. Thanks to You Li for his remarks on English mistakes.

\bibliographystyle{abbrv}
\bibliography{sigproc}  
%
%
\end{document}